\documentclass[10pt,twocolumn,letterpaper]{article}

\usepackage{cvpr}              
\usepackage{cuted}

\usepackage{multirow}
\usepackage{lipsum}
\usepackage[table]{xcolor}
\usepackage{amsmath}
\usepackage{pifont}
\usepackage[accsupp]{axessibility}
\newcommand{\cmark}{\ding{51}}%
\newcommand{\xmark}{\ding{55}}%
\definecolor{elvisrow}{RGB}{235,245,255} 

\definecolor{cvprblue}{rgb}{0.21,0.49,0.74}
\usepackage[pagebackref,breaklinks,colorlinks,allcolors=cvprblue]{hyperref}

\def\paperID{39714} 
\def\confName{CVPR}
\def\confYear{2026}

\title{ELVIS: Enhance Low-Light for Video Instance Segmentation in the Dark}

\author{
Joanne Lin \quad Ruirui Lin \quad Yini Li \quad David Bull \quad Nantheera Anantrasirichai \\
Visual Information Laboratory, University of Bristol \\
{\tt\small \{joanne.lin, r.lin, ub24017, dave.bull, n.anantrasirichai\}@bristol.ac.uk}
}

\begin{document}
\maketitle

\begin{strip}
    \centering
    \vspace{-0.75cm}
    \includegraphics[width=1.\textwidth]{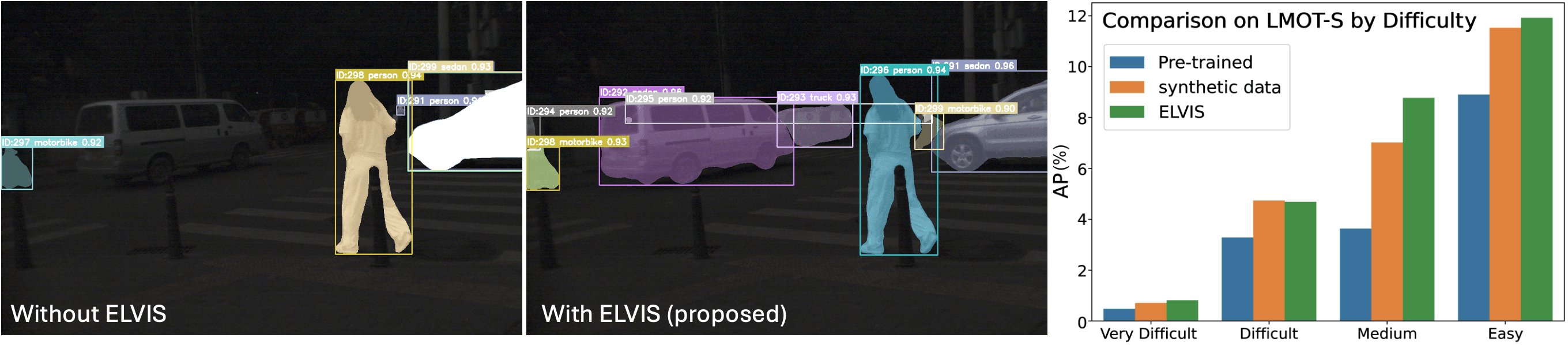}
    \captionof{figure}{\textbf{Video Instance Segmentation (VIS) in real low-light conditions.} VIS results using (left) GenVIS~\cite{heo2023genvis} and (center) GenVIS+ELVIS. (Right) Performance comparison of pre-trained VIS model, trained on synthetic data, and ELVIS across various difficulties.}
    \label{fig:hero}
\end{strip}

\begin{abstract}
Video instance segmentation (VIS) for low-light content remains highly challenging for both humans and machines alike, due to noise, blur and other adverse conditions.
The lack of large-scale annotated datasets and the limitations of current synthetic pipelines, particularly in modeling temporal degradations, further hinder progress. Moreover, existing VIS methods are not robust to the degradations found in low-light videos and, consequently, perform poorly even after finetuning.
In this paper, we introduce \textbf{ELVIS} (\textbf{E}nhance \textbf{L}ow-Light for \textbf{V}ideo \textbf{I}nstance \textbf{S}egmentation), a framework that enables domain adaptation of state-of-the-art VIS models to low-light scenarios. ELVIS is comprised of an unsupervised synthetic low-light video pipeline that models both spatial and temporal degradations, a calibration-free degradation profile estimation network (VDP-Net) and an enhancement decoder head that disentangles degradations from content features.
ELVIS improves performances by up to \textbf{+3.7AP} on the synthetic low-light YouTube-VIS 2019 dataset and beats two-stage baselines by at least \textbf{+2.8AP} on real low-light videos.
Code and dataset available at: \href{https://joannelin168.github.io/research/ELVIS}{https://joannelin168.github.io/research/ELVIS}
\end{abstract}    
\vspace{-2mm}
\section{Introduction}

Video instance segmentation (VIS) is a challenging yet important task in computer vision; with the goal of not only detecting, classifying and tracking an object, but also determining the pixel mask of the object.
This enables a fine-grained understanding of the object's state and motion. The task becomes even more complex under low-light conditions, which are common in applications including autonomous driving, wildlife conservation, surveillance, post-production and robotics.

Few datasets exist specifically for low-light data, reflecting the limited amount of research in this field. One contributing factor is that degradations caused by low-light conditions pose challenges for both human and automated annotation, as creating high-quality annotations for downstream tasks requires considerable effort. In particular, only a limited number of publicly available benchmarks~\cite{ye2022darktrack,lee2023lowlighthumanpose,liu2024ntvot211,wang2024lmot,Li2024LLEVOS} exist for downstream tasks in low-light conditions, and none of them are specifically designed for VIS applications.

Due to the scarcity of adequate training data, two main strategies are commonly employed to improve the performance of downstream tasks on low-light images and videos. One approach is to apply a pre-trained low-light enhancement model as a pre-processing step~\cite{wang2021sdsd, Cai2023retinexformer, lin2024bvi, bai2025retinexmamba} (commonly referred to as a two-stage approach); however, low-light \textit{video} enhancement itself is still in its infancy.
The other approach is to synthesize appropriate low-light conditions by modifying existing datasets for training~\cite{cui2021maet, Yi2024trackinglowlight, Li2024LLEVOS, lin2025segmlowlight}; these are commonly used to enable an end-to-end approach for downstream tasks in the dark.
Most existing synthetic approaches have been developed primarily for \textit{images}, reflecting the greater development and maturity of low-light image research. However, these methods often do not account for the blur degradations present in low-light videos, which result from the longer shutter speeds required to capture the content.

Furthermore, existing downstream methods are not designed to be robust to the degradations present in such conditions. Some works~\cite{cui2021maet, du2024zsdaobjectdetection, lin2025segmlowlight} focus on modifying architectures to improve robustness to degradations, others~\cite{Li2024LLEVOS, yao2025llevss, baek2025LLVOSEvents} leverage additional data (such as RAW data and events) for richer scene context, and~\cite{chen2023lis} combine both strategies. None of these, however, explore domain adaptation for VIS tasks in the dark.

To address these limitations, we propose a new synthetic low-light video pipeline and a domain adaptation framework (ELVIS) for improving the performance of existing VIS methods on low-light videos without the need for alternative data. An example of ELVIS’s capabilities is shown in~\cref{fig:hero}. Our contributions are three-fold:
\begin{itemize}
    \item We propose \textbf{ELVIS}, the \textbf{first} low-light Video Instance Segmentation (VIS) framework, which integrates an enhancement decoder into existing VIS architectures to disentangle degradation from scene content, thereby boosting performance in low-light videos.
    \item We develop the \textbf{first} physics-based degradation model for synthesizing \textbf{low-light videos}, incorporating illumination adjustment, blur degradations, and camera noise.
    \item We propose a novel Video Degradation Profiler Network (VDP-Net), trained in an \textbf{unsupervised manner}, to accurately estimate degradation profiles, enabling calibration-free synthesis of realistic low-light videos.
\end{itemize}
\section{Related Work}
\begin{figure*}[!ht]
    \includegraphics[width=\linewidth]{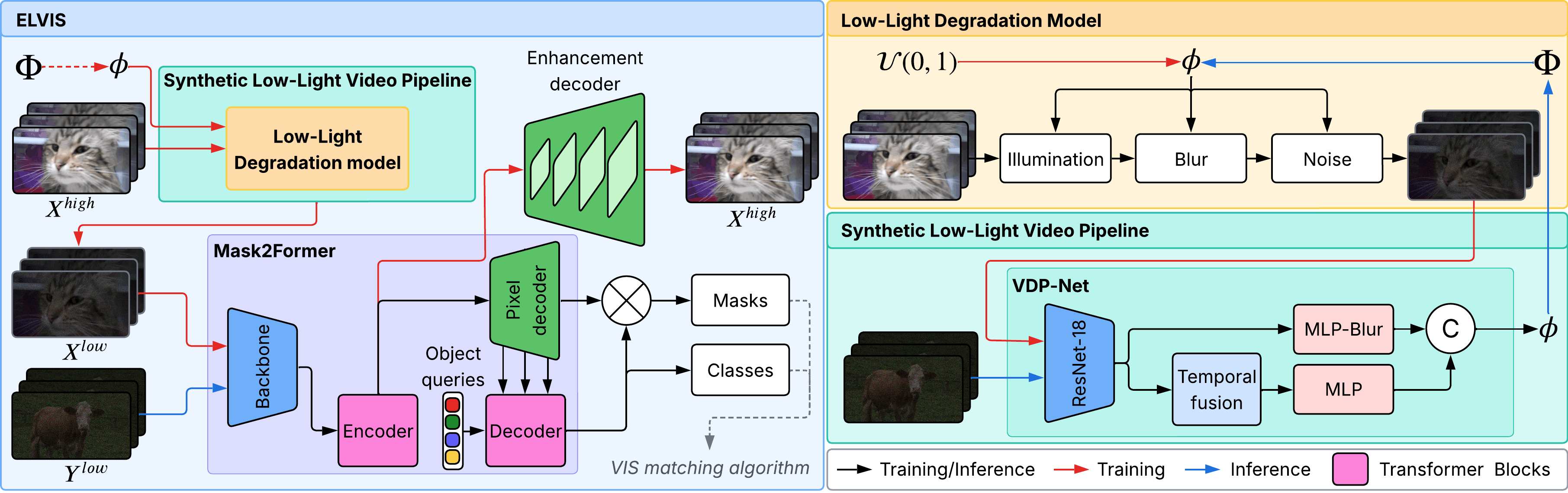}
    \caption{Overview of the proposed ELVIS framework, which consists of two main components: (i) the unsupervised synthetic low-light pipeline and (ii) the augmented instance segmentation module. The synthetic low-light video pipeline (green panel) degrades clean videos $X^{high}$ using the low-light degradation model (yellow panel) and degradation parameters $\phi$ estimated by VDP-Net.}
    \label{fig:diagram}
\end{figure*}

\subsection{Synthetic Low-Light Pipelines}

\noindent \textbf{Physics-based pipelines.}
One popular synthetic low-light pipeline from Lv~\etal~\cite{lv2021agllnet} considers the noise as a result of demosaicing in addition to read and shot noise. They make use of the heteroscedastic noise model~\cite{foi2008noiseapprox}, which uses a Gaussian distribution as an approximation for combining read and shot noise (often modeled separately using a Gaussian and Poisson distribution). Although this is a sufficient approximation for brightly-lit conditions, it fails to accurately model read and shot noise in low-light conditions~\cite{wang2021extremenoise}. Many works exploring low-light downstream tasks~\cite{zhang2021stablellve, Li2024LLEVOS, lin2025segmlowlight} make use of this synthetic pipeline for their research, due to its simplicity and versatility.

Wei~\etal~\cite{wei2021eld} propose a RAW noise pipeline which involves meticulous experimentation with camera-metadata for collecting the noise parameters for their proposed noise model. They calculate the noise profiles for 5 different cameras using statistical methods. Their synthetic pipeline models the following noise types using various statistical distributions: read noise, shot noise, quantization noise, and banding noise. This work has been used by Chen~\etal~\cite{chen2023lis} for instance segmentation in RAW low-light images.

\vspace{2mm}
\noindent \textbf{Deep-learning pipelines.}
Building upon physics-based methods, Monakhova~\etal~\cite{monakhova2022starlight} designed a Wasserstein GAN~\cite{arjovsky2017wgan} to synthesize bursts of noisy RAW frames from a clean still RAW image. The generator learns to first estimate the degradation parameters of their Submillilux dataset, followed by a U-Net~\cite{ronneberge2015unet} to further model complex degradations. While effective for their dataset, the method lacks generalizability and requires re-training for other use cases. Meanwhile, Zhou~\etal~\cite{zhou2022lednet} explored a multi-step approach to synthesizing low-light sRGB videos which considered illumination, blur and noise degradations. However, their method still has limitations as they incorporate supervised methods~\cite{Niklaus2017sepconv,zamir2020cycleisp} for synthesizing the degradations.

\subsection{Calibration-free Degradation Estimation}
Several works have explored estimating degradations within images and videos, particularly noise, for calibration-free, physics-based synthesis. 
Jin~\etal~\cite{jin2023led}, Zou~\etal~\cite{zou2025rawnoiseetimation}, and Lin~\etal~\cite{lin2025den} adopt approaches similar to ours by leveraging physics-based statistical distributions to synthesize noise, though each have notable limitations.~\cite{jin2023led, zou2025rawnoiseetimation} rely on the manually collected degradation parameters from~\cite{wei2021eld} to train their noise estimation model. This requires image-capture metadata such as camera model and ISO, which are typically unavailable in publicly released datasets used for downstream tasks. Lin~\etal address this issue by training a U-Net~\cite{ronneberge2015unet} to estimate degradation parameters across diverse noise distributions; however, their method is tailored towards extreme low-light noise.

\subsection{Domain adaptation via synthetic data}
Many works~\cite{zhang2021stablellve, Li2024LLEVOS, yao2025llevss, lin2025segmlowlight} use pre-existing pipelines for synthesizing their training data for downstream tasks under low-light conditions, while others propose their own synthetic pipelines~\cite{luo2023similarityminmax, cui2021maet}.
These methods typically prioritize efficiency over photorealism, enabling on-the-fly synthesis and increased training diversity.
Cui~\etal~\cite{cui2021maet} model in-camera degradations by `unprocessing' sRGB images to RAW using~\cite{brooks2019unprocessing}, injecting read and shot noise (modeled with Gaussian and Poisson distributions respectively), and reconstructing images via a simple ISP. They incorporate this low-light pipeline within their training objective, by jointly learning to estimate the relevant degradation parameters alongside object detection.
Du~\etal~\cite{du2024zsdaobjectdetection} adopt this pipeline but enforce domain adaptation by jointly predicting detections and reflectance components, inspired by Retinex theory~\cite{land1977retinex}. 
Similarly, Chen~\etal~\cite{chen2023lis} synthesize low-light RAW images by unprocessing to RAW~\cite{brooks2019unprocessing} and applying the noise model of Wei~\etal~\cite{wei2021eld}.
\section{Methodology}

An overview of the proposed Enhance Low-Light for Video Instance Segmentation (ELVIS) framework is shown in~\cref{fig:diagram}. The ELVIS framework introduces two novel components: (i) an unsupervised Synthetic Low-Light Pipeline, detailed in~\cref{sec:pipeline}, and (ii) an Enhancement Decoder Head within the VIS network, which reconstructs the original video and thereby improves the perceptual capability of VIS methods under low-light conditions, as described in~\cref{ssec:method}. Within the synthetic low-light pipeline, we introduce the Video Degradation Profiler Network (VDP-Net)~(\cref{ssec:VDPNet}), which generates degradation parameters used to synthesize low-light videos during ELVIS training.

We first train VDP-Net by uniformly sampling degradation parameters $\phi$. Once fully trained, VDP-Net can estimate $\phi$ for any given low-light input. Using these estimated parameters, synthetic low-light videos are then generated through the proposed pipeline, which comprises illumination adjustment, blur, and noise modeling. To train ELVIS, we collect a set of degradation profiles from public real low-light video datasets and synthesize low-light videos on the fly using the proposed low-light degradation model.

\subsection{Synthetic Low-Light Video Pipeline}
\label{sec:pipeline}

We propose a new unsupervised synthetic low-light pipeline for generating videos to facilitate low-light domain adaptation of downstream tasks. This pipeline introduces a novel low-light degradation model that synthesizes low-light videos given a vector of parameters $\phi$, estimated by a calibration-free VDP-Net.

Our synthetic low-light video pipeline aims to map videos from normal-light to low-light conditions in a fully unsupervised manner. We denote a video $X$ (with $T$ frames, $C$ channels, height $H$, and width $W$) captured under normal lighting as $X^{high}$, and its corresponding low-light version as $X^{low}$. The pipeline emulates real video acquisition in digital cameras by modeling exposure, blur, and noise, as illustrated in the yellow panel of~\cref{fig:diagram}.

\subsubsection{Illumination adjustment}
In the first step of the synthesis procedure, we reduce the illumination of the videos to simulate low-light conditions in the scene. For a linear image representation (e.g., raw sensor data), pixel intensity is directly proportional to scene exposure. Since most images are stored in the sRGB color space, we first convert them to the XYZ color space to ensure linearity. The brightness of a frame $\mathbf{x}$ can then be reduced using the transformation $\mathbf{x}' = \alpha \mathbf{x}$. In photography, a one-stop adjustment in shutter speed, aperture, or ISO corresponds to a doubling or halving of scene exposure; therefore, we rewrite the illumination adjustment function as:
\begin{equation}
    X' = 2^{\epsilon} X,
\end{equation}
where $X'$ is the darkened video output, $\epsilon$ is the exposure adjustment value and $X$ is the input video.


\subsubsection{Blur degradation with multivariate Gaussian}
We also incorporate blur degradations into our synthetic pipeline—an aspect often overlooked in other synthetic degradation frameworks—despite being a common artifact caused by long shutter speeds, a typical camera setting in low-light environments. Two types of blur are typically present in low-light videos: motion blur and defocus blur.

\begin{figure}[!t]
    \centering
    \includegraphics[width=\linewidth]{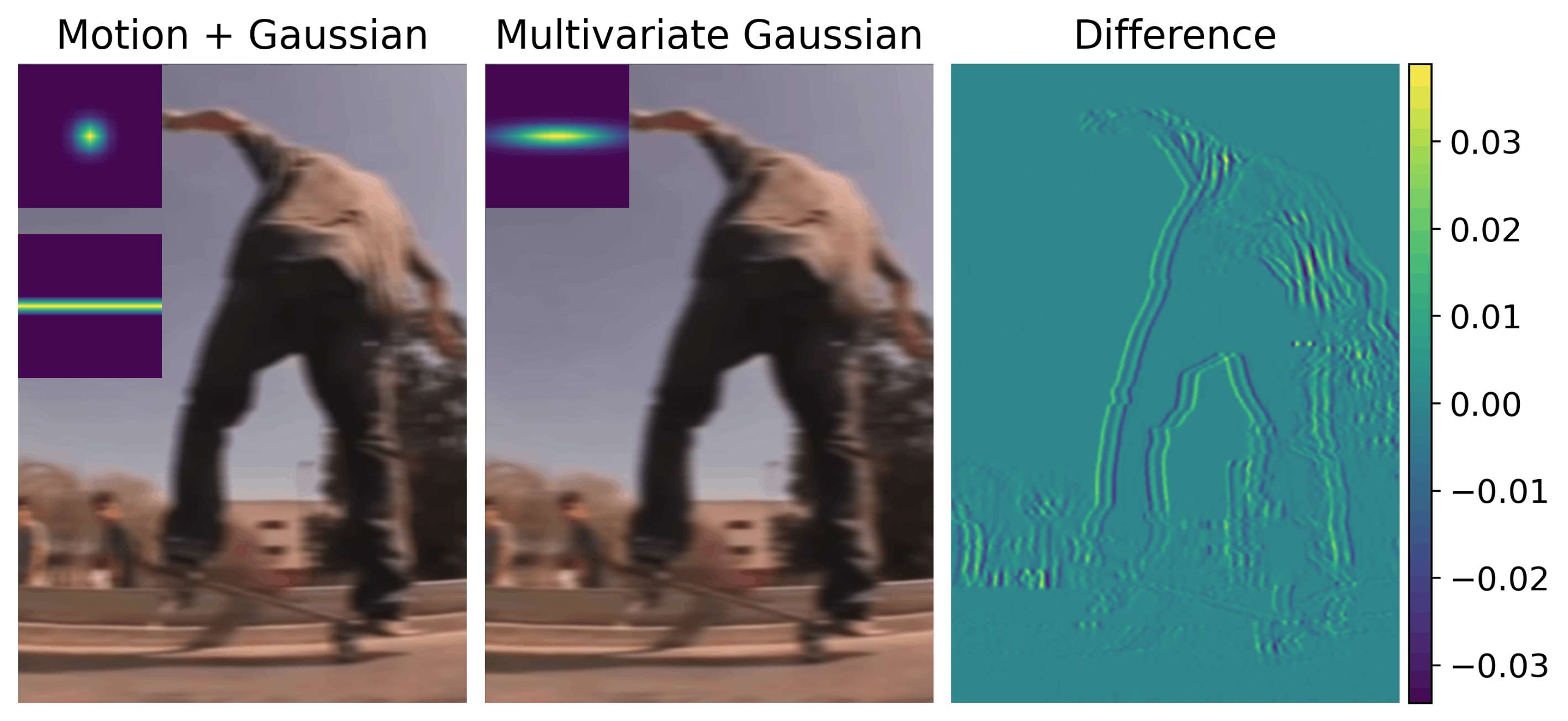}
    \caption{Visual comparison of the linear motion blur + Gaussian blur, the Multivariate Gaussian blur, and the difference between the two resulting images (intensities normalized to [-1, 1]). The blur kernels can be seen in the top-left of each blurred image.}
    \label{fig:blur_comparison}
\end{figure}

Traditionally, linear motion blur is modeled using a line-shaped point spread function (PSF) aligned with the motion direction, while defocus blur is modeled using a Gaussian PSF kernel. An alternative approach to simulating motion blur involves averaging consecutive frames~\cite{nah2017dynamicdeblur, su2017handheldvideodeblur, li2021ARVOdeblur, zhou2022lednet}. However, this method offers limited control over blur magnitude, requires small inter-frame motion (thus necessitating high frame rates to avoid ghosting artifacts), discards boundary frames during averaging, and is computationally expensive.
Since the effects of motion and defocus blur are difficult to separate, we model their combined effect using a multivariate Gaussian distribution~\cite{pretto2009bivariategaussianblur}. This approach not only addresses the limitations of frame averaging but also enables VDP-Net to better learn blur degradations with minimal visual discrepancies. As illustrated in~\cref{fig:blur_comparison}, the maximum intensity difference between the traditional method and our approach is less than 4\%, indicating very close visual similarity. Our blur model is a deliberate design choice; balancing realism, efficiency, and unsupervised learnability.

We approximate the joint blur kernel $H$, using three parameters: $\sigma_{H_x}, \sigma_{H_y}, \theta_H$, as shown in~\cref{eqn:blur_covar,eqn:blur_kernel}:
\begin{equation}
    \Sigma = R
    \begin{bmatrix}
    \sigma_{H_x}^2 & 0 \\
    0 & \sigma_{H_y}^2
    \end{bmatrix}
    R^T, \:
    R =
    \begin{bmatrix}
    \cos\theta_H & -\sin\theta_H \\
    \sin\theta_H & \cos\theta_H
    \end{bmatrix},
\label{eqn:blur_covar}
\end{equation}
\begin{equation}
    H[i,j] = \frac{
    \exp\Big(-\frac{1}{2} \mathbf{r}_{ij}^\top \Sigma^{-1} \mathbf{r}_{ij} \Big)
    }{
    \sum_{i,j} \exp\Big(-\frac{1}{2} \mathbf{r}_{ij}^\top \Sigma^{-1} \mathbf{r}_{ij} \Big)
    }, 
\label{eqn:blur_kernel}
\end{equation}
where $i, j$ refer to the indices of the kernel, $\Sigma$ is the covariance matrix to determine the spread of the multivariate Gaussian blur kernel, $R$ is the rotation matrix that orientates the 2D Gaussian PSF, and $\mathbf{r}_{ij}$ is the position vector relative to the kernel center. The kernel is normalized such that its sum equals one.

We constrain $\theta_H$ to $[0,\frac{\pi}{2}]$ when $\sigma_{H_x} > \sigma_{H_y}$; otherwise, the kernel orientation is assigned to $[\frac{\pi}{2}, \pi]$. Overall, this ensures that the motion-blur kernel angle lies within $[0,\pi]$, because the motion blur kernel is bidirectional (i.e. orientations separated by $\pi$ radians are equivalent). When $\sigma_{H_x}$ equals $\sigma_{H_y}$, the video exhibits only defocus blur.

\subsubsection{Physics-based noise}
Our low-light degradation model focuses on physics-based noise arising from the limitations of the camera sensor. We do not consider spatially correlated artifacts introduced or amplified by post-processing steps such as compression, demosaicing, or in-camera denoising, as these are unique to each camera's image signal processor (ISP).

i) \textbf{Read noise} arises from electrical noise present in the signal readout process of the camera sensor. It is a combination of a variety of noise types \cite{wei2021eld} that can generally be modeled using a Gaussian distribution. 
\begin{equation}
     N_r \sim \mathcal{N}(0, \sigma_r^2), \quad N_r \in \mathbb{R}^{T \times C \times H \times W},
     \label{eqn:noise_read}
\end{equation}
where $N_r$ is the read noise map, and $\sigma_r^2$ is the variance for the Gaussian distribution.

ii) \textbf{Shot noise} is found in low-light capture; resulting from the limited photons hitting the camera sensor from the lack of light in the scene. This behavior can be modeled using a Poisson distribution $\mathcal{P}(\cdot)$
\begin{equation}
     (X + N_s) \sim \mathcal{P}(\frac{X}{K})K, \quad N_r \in \mathbb{R}^{T \times C \times H \times W},
     \label{eqn:noise_shot}
\end{equation}
where $N_s$ is the shot noise map, and $K$ denotes the overall system gain.

iii) \textbf{Quantization noise} arises during the process of discretizing the analog signal into a digital signal. This conversion often introduces visual artifacts in the form of grainy or blocky images, which is especially noticeable in images with low bit depth. We model this using a uniform distribution 
\begin{equation}
     N_q \sim \mathcal{U}(0, \lambda_q), \quad N_q \in \mathbb{R}^{T\times C \times H \times W},
     \label{eqn:noise_quantization}
\end{equation}
where $N_q$ is the quantization noise map, and $\lambda_q$ is the upper bound for the quantization noise interval.

iv) \textbf{Banding noise}—camera-specific horizontal or vertical lines prominent at high ISO—can be modeled as zero-mean Gaussian noise~\cite{monakhova2022starlight, wei2021eld} with standard deviation $\sigma_b$ controlling band variability. Unlike prior works that model a single orientation, we model \textit{both} for greater flexibility. The banding noise is modeled as
\begin{equation}
    N_b \sim 
        \begin{cases}
            \mathcal{N}(0, \sigma_b^2), & N_b \in \mathbb{R}^{T\times C \times 1 \times W}, \quad \text{if } \theta_b = 0 \\
            \mathcal{N}(0, \sigma_b^2), & N_b \in \mathbb{R}^{T\times C \times H \times 1}, \quad \text{if } \theta_b = 1
        \end{cases}
    \label{eqn:noise_banding}
\end{equation}
where $N_b$ is the banding noise map and $\theta_b$ determines the orientation of the bands.

\subsubsection{Final Low-Light Degradation Model}
Combining above degradations, our synthesis function can be expressed as follows:
\begin{equation}
    X^{low} = Deg(X^{high}, \phi) = H *(2^\epsilon X^{high}) + N 
    \label{eqn:synthesis}
\end{equation}
where $*$ denotes applying a 2D convolution of the kernel $H$ onto video $X^{high}$, and $N$ is total noise value derived from~\cref{eqn:noise_read,eqn:noise_shot,eqn:noise_quantization,eqn:noise_banding}; denoted in a similar fashion to ~\cite{wei2021eld, wang2021extremenoise, monakhova2022starlight, cao2023ISOdependentnoise}. We define $\phi$ as a vector of parameters, which are used to apply the degradations onto the video. As such, $\phi$ can be thought of as a degradation profile.
\begin{equation}
    \phi = \{ \epsilon, \sigma_r, K, \lambda_q, \sigma_b, \theta_b, \sigma_{H_x}, \sigma_{H_y}, \theta_H \}.
    \label{eq:phi}
\end{equation}

\subsection{Video Degradation Profiler Network}
\label{ssec:VDPNet}
To estimate the degradation parameters $\phi$, we employ our proposed VDP-Net, which efficiently extracts accurate degradation profiles from real low-light videos without the need for manual calibration. VDP-Net is a crucial component of the ELVIS framework as it allows for the synthetic pipeline to sample from realistic degradation profiles, instead of potentially unrealistic profiles due to random sampling.
VDP-Net consists of a lightweight feature extractor, a temporal fusion convolutional block, and two multi-layer perceptrons (MLPs) for predicting  $\phi$.

For the backbone, we use a lightweight pre-trained ResNet-18~\cite{he2016resnet}, as it enables faster convergence. Because exposure and noise in low-light video clips are global degradations, whereas blur is a local degradation whose characteristics can vary significantly (particularly in the case of motion blur), we design VDP-Net with two separate prediction heads. The temporal fusion module aggregates features from the backbone along the temporal dimension and passes them to an MLP head to output exposure and noise degradation parameters. The temporal fusion block consists of an average pooling layer, a one-dimensional convolutional layer, batch normalization, and a ReLU activation function.

\vspace{2mm}
\noindent \textbf{Unsupervised training strategy.}
Since real low-light videos lack ground truth degradation parameters, we adopt an unsupervised training strategy in which random values of $\phi$ are uniformly sampled to generate degraded inputs $X^B$ for the network. We first consulted domain experts to determine realistic upper bounds for the degradation levels observed in real low-light conditions. During training, values of $\phi$ within these bounds are passed to our low-light degradation model, together with a clean normal-light video $X^A$, to synthesize the corresponding low-light video $X^B$. Uniform sampling ensures that the network learns the full range of degradations without bias.

\vspace{2mm}
\noindent \textbf{Loss function.}
We find that using only the L1 loss between the ground truth $\phi$ and the VDP-Net prediction $\phi'$ is insufficient, particularly for the motion blur parameters. This is because the L1 loss does not account for the cyclical nature of the blur angle. To address this, we introduce a cosine angular loss (CA loss) term  that effectively handles this periodic behavior. The overall loss function is defined as
\begin{equation}
\mathcal{L}_{\text{total}} = \lambda_1 \lVert\phi - \phi'\rVert_1 + \lambda_2 \left(1 - \cos(|\theta_H - \theta_H'|)\right),
\end{equation}
where $\lambda_1$ and $\lambda_2$ are weighting coefficients for each loss component.
\subsection{Video Instance Segmentation in ELVIS}
\label{ssec:method}

\subsubsection{Enhancement decoder integration}

Existing VIS methods~\cite{huang2022minvis, heo2023genvis, zhang2025dvis++} are not specifically designed to handle degraded videos and therefore perform poorly, even when re-trained on synthetic low-light data, as shown in~\cref{fig:visual_comparison_lmots_frames_segms}. To address this limitation, our framework, ELVIS, introduces an enhancement decoder head within the segmentation module of the network. We integrate this decoder into the Mask2Former~\cite{cheng2021mask2former} architecture, which serves as the foundation for many state-of-the-art VIS models.

The proposed enhancement decoder employs a multi-scale deformable attention pixel decoder~\cite{zhu2021deformable} consisting of ten transformer decoder layers combined with bilinear upsampling layers to reconstruct normal-light frames. We train the enhanced VIS network by adding an L1 loss between the clean normal-light videos and the decoder’s reconstructed outputs, as a simple-yet-effective loss function. This additional head maps the latent features of Mask2Former to normal-light representations, guiding the network to disentangle scene content from degradations present in low-light videos.

\subsubsection{Training strategy with synthetic low-light data}
For VIS training, we first generate a large set of degradation parameters $\Phi$ estimated by our VDP-Net across multiple real low-light video datasets (SDSD~\cite{wang2021sdsd}, DID~\cite{fu2023did}, BVI-RLV~\cite{lin2024bvi}, and LMOT~\cite{wang2024lmot}). We selected these datasets because they cover a broad range of degradation levels, most of which can be seen in~\cref{fig:visual_comparison_synth_crops}. During VIS training, values of $\phi$ are randomly sampled from these pre-generated parameters $\Phi$ and applied to the input training videos. Collecting $\Phi$ in advance, rather than generating them uniformly at random as done when training VDP-Net, enables the VIS model to learn under realistic degradation conditions.
\section{Experimental Results}
\begin{table*}
    \centering
        \small
         \caption{Evaluation on the synthetic YouTube-VIS 2019~\cite{yang2019ytvis19} validation set using several SOTA online VIS methods and backbone networks, each trained on synthetic low-light videos generated with our synthetic pipeline. \textbf{Bold} denotes the better performances.}
    \begin{tabular}{l|c|c|ccccccccc}
    \toprule
     Method & Backbone & ELVIS & AP & AP$_{50}$ & AP$_{75}$ & AP$_{S}$ & AP$_{M}$ & AP$_L$ & AR$_{1}$ & AR$_{10}$ \\
    \midrule

    MinVIS~\cite{huang2022minvis} & ResNet-50 & \xmark & 36.4 & \textbf{57.3} & 36.4 & 13.3 & 30.6 & \textbf{49.4} & 36.5 & 44.4 \\
    GenVIS~\cite{heo2023genvis} & ResNet-50 & \xmark & 39.1 & 58.4 & 42.7 & 16.2 & 34.8 & \textbf{55.2} & 40.3 & 48.4 \\
    DVIS++~\cite{zhang2025dvis++} & ResNet-50 & \xmark & 38.8 & 59.9 & 42.8 & 23.8 & 38.5 & 51.4 & 39.5 & 49.6 \\
    
    \rowcolor{elvisrow}
    MinVIS~\cite{huang2022minvis} & ResNet-50 & \cmark & \textbf{37.2} & 57.0 & \textbf{39.6} & \textbf{15.1} & \textbf{34.5} & 30.9 & \textbf{37.8} & \textbf{45.7} \\
    \rowcolor{elvisrow}
    GenVIS~\cite{heo2023genvis} & ResNet-50 & \cmark & \textbf{41.0} & \textbf{59.8} & \textbf{46.2} & \textbf{18.5} & \textbf{38.0} & 54.2 & \textbf{42.0} & \textbf{51.2} \\
    \rowcolor{elvisrow}
    DVIS++~\cite{zhang2025dvis++} & ResNet-50 & \cmark & \textbf{42.5} & \textbf{63.8} & \textbf{46.6} & \textbf{27.6} & \textbf{42.5} & \textbf{54.3} & \textbf{41.7} & \textbf{51.9} \\
    
    \midrule
    MinVIS~\cite{huang2022minvis} & SWIN-L & \xmark & 51.8 & 73.8 & 57.9 & 28.0 & 43.0 & 69.8 & 46.7 & 57.5 \\
    GenVIS~\cite{heo2023genvis} & SWIN-L & \xmark & 53.7 & 74.8 & 58.7 & \textbf{27.6} & 44.9 & 71.0 & 49.1 & 59.6 \\
    DVIS++~\cite{zhang2025dvis++} & ViT-L & \xmark & 55.2 & 77.2 & 62.1 & \textbf{29.5} & 51.6 & 70.5 & 49.7 & 61.5 \\
    
    \rowcolor{elvisrow}
    MinVIS~\cite{huang2022minvis} & SWIN-L & \cmark & \textbf{54.2} & \textbf{78.3} & \textbf{61.6} & \textbf{28.4} & \textbf{47.6} & \textbf{71.7} & \textbf{49.8} & \textbf{60.1} \\
    \rowcolor{elvisrow}
    GenVIS~\cite{heo2023genvis} & SWIN-L & \cmark & \textbf{55.3} & \textbf{79.3} & \textbf{61.1} & 26.6 & \textbf{49.6} & \textbf{72.3} & \textbf{49.5} & \textbf{61.2} \\
    \rowcolor{elvisrow}
    DVIS++~\cite{zhang2025dvis++} & ViT-L & \cmark & \textbf{56.9} & \textbf{78.7} & \textbf{65.3} & 25.5 & \textbf{54.3} & \textbf{74.8} & \textbf{50.3} & \textbf{62.7} \\
    \bottomrule
    \end{tabular}
    \vspace{-10pt}
    \label{tab:results_segm_synth}
\end{table*}


\begin{table}
 \centering
 \caption{Evaluation on LMOT-S across several SOTA online VIS methods, trained on the synthetic low-light YouTube-VIS 2019~\cite{yang2019ytvis19} videos. \textbf{Bold} denotes the better performances.}
    \setlength{\tabcolsep}{3pt}
    \resizebox{1.0\linewidth}{!}{
    \begin{tabular}{c|l|c|cccccc}
    \toprule
    \multicolumn{2}{l|}{Method} & ELVIS & AP & AP$_{50}$ & AP$_{75}$ & AR$_{1}$ & AR$_{10}$ & AR$_{100}$\\
    \midrule
    
    \multirow{6}{*}{\rotatebox{90}{ResNet-50}}
    & MinVIS~\cite{huang2022minvis} & \xmark & 4.1 & 9.5 & 2.5 & 3.0 & 4.7 & 4.8 \\
    & GenVIS~\cite{heo2023genvis} & \xmark & 6.6 & 14.5 & \textbf{5.2} & \textbf{5.3} & 9.8 & 11.7 \\
    & DVIS++~\cite{zhang2025dvis++} & \xmark & 7.0 & 15.9 & \textbf{4.4} & 4.7 & 9.3 & \textbf{10.3} \\
    & \cellcolor{elvisrow}MinVIS~\cite{huang2022minvis} & \cellcolor{elvisrow}\cmark & \cellcolor{elvisrow}\textbf{4.9} & \cellcolor{elvisrow}\textbf{11.7} & \cellcolor{elvisrow}\textbf{3.2} & \cellcolor{elvisrow}\textbf{3.7} & \cellcolor{elvisrow}\textbf{5.5} & \cellcolor{elvisrow}\textbf{5.6} \\
    & \cellcolor{elvisrow}GenVIS~\cite{heo2023genvis} & \cellcolor{elvisrow}\cmark & \cellcolor{elvisrow}\textbf{6.7} & \cellcolor{elvisrow}\textbf{15.5} & \cellcolor{elvisrow}4.4 & \cellcolor{elvisrow}\textbf{5.3} & \cellcolor{elvisrow}\textbf{12.1} & \cellcolor{elvisrow}\textbf{14.0}\\
    & \cellcolor{elvisrow}DVIS++~\cite{zhang2025dvis++} & \cellcolor{elvisrow}\cmark & \cellcolor{elvisrow}\textbf{7.3} & \cellcolor{elvisrow}\textbf{16.7} & \cellcolor{elvisrow}4.3 & \cellcolor{elvisrow}\textbf{4.8} & \cellcolor{elvisrow}\textbf{9.4} & \cellcolor{elvisrow}\textbf{10.3} \\

    \midrule
    
    \multirow{2}{*}{\rotatebox{90}{ViT-L}}
    & DVIS++~\cite{zhang2025dvis++} & \xmark & 10.0 & 21.4 & 7.6 & 5.8 & 13.1 & \textbf{17.5} \\
    & \cellcolor{elvisrow}DVIS++~\cite{zhang2025dvis++} & \cellcolor{elvisrow}\cmark & \cellcolor{elvisrow}\textbf{10.5} & \cellcolor{elvisrow}\textbf{22.6} & \cellcolor{elvisrow}\textbf{8.3} & \cellcolor{elvisrow}\textbf{6.0} & \cellcolor{elvisrow}\textbf{14.5} & \cellcolor{elvisrow}16.8 \\

    \bottomrule
    \end{tabular}}
    \label{tab:results_segm_lmots}
\end{table}


\begin{table}
    \centering
       \caption{Evaluation against two-stage baselines on ELVIS-S and LMOT-S. \textbf{Bold} indicates the best performances.}
    \small
    \begin{tabular}{l|ccccc}
        \toprule
        Method & AP & AP$_{50}$ & AP$_{75}$ & AR$_{1}$ & AR$_{10}$ \\
        \midrule
        \multicolumn{6}{l}{\textit{ELVIS-S}} \\
        \midrule
        SDSD-Net~\cite{wang2021sdsd} & 46.7 & 96.0 & 26.6 & 41.9 & 47.8 \\
        StableLLVE~\cite{zhang2021stablellve} & 57.3 & \textbf{96.8} & \textbf{47.7} & 47.2 & 57.8 \\
        DarkIR~\cite{feijoo2025DarkIR} & 55.9 & \textbf{96.8} & 45.5 & 47.2 & 56.6 \\
        ELVIS & \textbf{58.0} & \textbf{96.8} & 47.6 & \textbf{48.1} & \textbf{58.8} \\
        
        \midrule
        \multicolumn{6}{l}{\textit{LMOT-S}} \\
        \midrule
        SDSD-Net~\cite{wang2021sdsd} & 2.5 & 6.4 & 1.8 & 2.5 & 5.2\\
        StableLLVE~\cite{zhang2021stablellve} & 3.9 & 8.3 & 3.0 & 3.7 & 7.6\\
        DarkIR~\cite{feijoo2025DarkIR} & 3.8 & 8.7 & 3.0 & 4.1 & 7.7 \\
        ELVIS & \textbf{6.7} & \textbf{15.5} & \textbf{4.4} & \textbf{5.3} & \textbf{12.1} \\
        
        \bottomrule
    \end{tabular}
    \vspace{-5pt}
    \label{tab:results_comparisons_twostage}
\end{table}

\begin{table}
    \centering
       \caption{Evaluation against unsupervised synthetic low-light pipelines on ELVIS-S and LMOT-S. \textbf{Bold} indicates the best performances.}
    \small
    \begin{tabular}{l|ccccc}
        \toprule
        Method & AP & AP$_{50}$ & AP$_{75}$ & AR$_{1}$ & AR$_{10}$ \\
        \midrule
        \multicolumn{6}{l}{\textit{ELVIS-S}} \\
        \midrule
        Lv \etal~\cite{lv2021agllnet} & 53.5 & 92.7 & \textbf{43.5} & 45.3 & 54.1 \\
        Cui \etal~\cite{cui2021maet} & 51.1 & \textbf{96.8} & 43.0 & 45.3 & 51.9 \\
        Lin \etal~\cite{lin2025den} & 52.8 & \textbf{96.8} & 40.8 & 45.0 & 53.4\\
        Ours (random) & 39.9 & 70.6 & 26.6 & 35.6 & 45.0 \\
        Ours & \textbf{54.5} & \textbf{96.8} & 43.0 & \textbf{46.6} & \textbf{55.3} \\ 
        
        \midrule
        \multicolumn{6}{l}{\textit{LMOT-S}} \\
        \midrule
        Lv \etal~\cite{lv2021agllnet} & 5.1 & 10.8 & 4.6 & 5.2 & 8.8 \\
        Cui \etal~\cite{cui2021maet} & 5.7 & 13.1 & 4.1 & 4.9 & 9.6 \\
        Lin \etal~\cite{lin2025den} & 6.3 & \textbf{15.0} & 4.2 & 5.2 & 8.5 \\
        Ours (random) & 4.7 & 12.1 & 2.5 & 4.0 & 6.8 \\
        Ours & \textbf{6.6} & 14.5 & \textbf{5.2} & \textbf{5.3} & \textbf{9.8}\\ 
        
        \bottomrule
    \end{tabular}
    \vspace{-5pt}
    \label{tab:results_comparisons_pipelines}
\end{table}

\subsection{Low-Light Video Instance Segmentation}

\noindent \textbf{Implementation Details.}
We selected state-of-the-art (SOTA) VIS methods which incorporate a Mask2Former~\cite{cheng2021mask2former} segmentation head. We trained these using ELVIS on the synthetic YouTube-VIS 2019~\cite{yang2019ytvis19} dataset by passing its videos into our synthetic low-light pipeline and sampling $\phi$ from the large degradation profile set $\Phi$. For fair comparison, the benchmark models were also re-trained on synthetic YouTube-VIS 2019.

We evaluate ELVIS on both synthetic and real low-light videos. For the synthetic data experiments, we applied our synthetic pipeline to the YouTube-VIS 2019 validation set (as described above) and evaluated performance using the provided metrics.
Due to the absence of publicly available benchmarks for real low-light video instance segmentation (VIS), we evaluate on two complementary datasets: a small manually-annotated dataset \textit{ELVIS-S} and a larger dataset with pseudo-ground truth \textit{LMOT-S}.
For ELVIS-S, we capture 5 real low-light videos using a RED V-Raptor camera. All frames are cropped to 1080p resolution, and 50 frames per video are densely annotated, resulting in 250 labeled frames in total.
For LMOT-S, we generate pseudo-ground truth instance masks on the sRGB validation split of the LMOT dataset~\cite{wang2024lmot} using the Segment Anything Model (SAM)~\cite{kirillov2023segmentanything} (refer to~\cref{ssec:suppmat_lmots_construction} for construction details). We also raise the number of detections to 100 when evaluating on LMOT-S, due to its high instance count. This larger dataset allows us to evaluate performance on data with more variance while avoiding the cost of manual annotation.

\begin{figure*}[ht]
\centering
\includegraphics[width=0.75\textwidth]{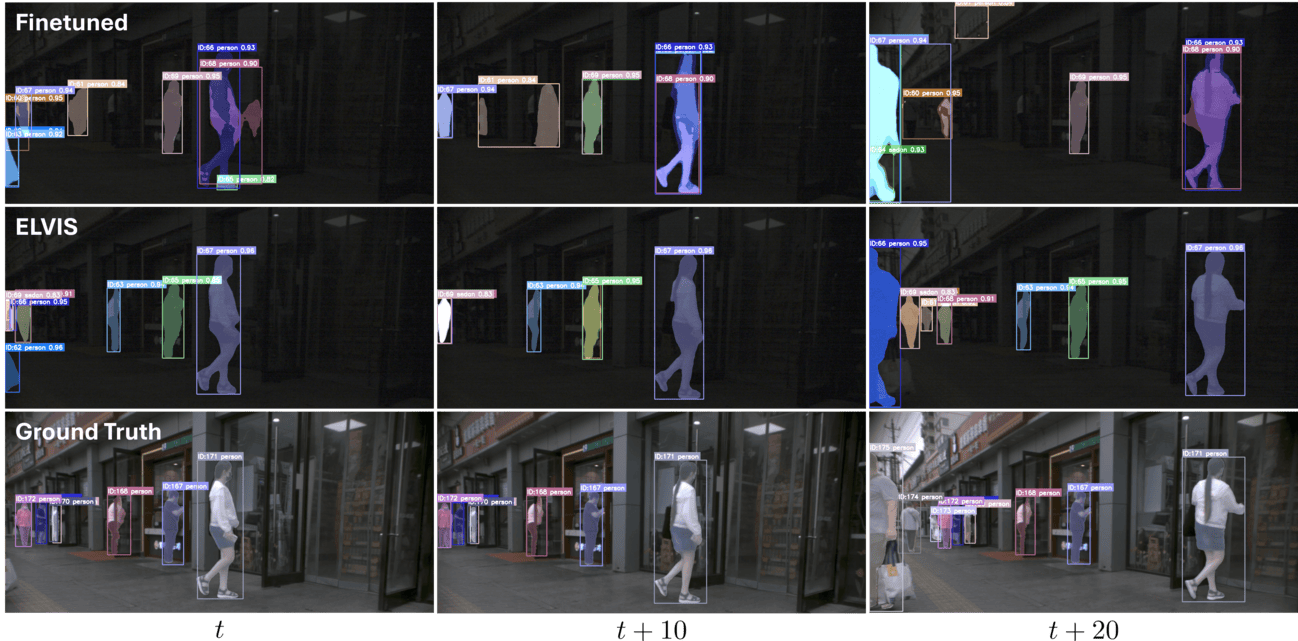}
\caption{Visual comparison of VIS results on the LMOT-S dataset using GenVIS R50~\cite{heo2023genvis, he2016resnet} finetuned on our synthetic data (top row) versus implementing our ELVIS framework (middle row), with ground truth (bottom row) for reference. The columns represent frames in the example video, sampled every 10 frames from time $t$, to show the tracking performances.}
\label{fig:visual_comparison_lmots_frames_segms}
\end{figure*}

\subsubsection{Evaluation on synthetic low-light data}
We select 3 SOTA VIS methods, MinVIS~\cite{huang2022minvis}, GenVIS~\cite{heo2023genvis} and DVIS++~\cite{zhang2025dvis++} to demonstrate the versatility of our proposed ELVIS framework; the results are shown in~\cref{tab:results_segm_synth}. For MinVIS and GenVIS, we allow the backbones to be finetuned, while for DVIS++, we first train its segmenter network and then we freeze it to train its tracker network. For each of these methods, we use the same training configurations as stated in the respective works.
We implement both a CNN backbone and a Transformer backbone to further validate the flexible application of ELVIS.
For the CNN backbone experiments we employ the popular ResNet-50~\cite{he2016resnet} (R50), while for the Transformer backbone experiments, we select the SWIN-L~\cite{Liu2021swin} backbone for MinVIS and GenVIS and ViT-L~\cite{dosovitskiy202ViT} backbone for DVIS++.
It is clear that ELVIS improves performances across all metrics;
we show significant improvements to AP$_{50}$ and AP$_{75}$ demonstrating that our method produces masks with higher Intersection-over-Union (IoU) accuracy. The improvements to AR show that ELVIS also reduces false negatives.

\subsubsection{Evaluation on real low-light data}
In our preliminary real low-light experiments, we verify on LMOT-S that ELVIS improves SOTA VIS performances (\cref{tab:results_segm_lmots}). We also compare against two-stage baselines (applying a low-light enhancement method followed by a pre-trained GenVIS R50) and GenVIS trained on other synthetic pipelines.
~\cref{tab:results_comparisons_twostage,tab:results_comparisons_pipelines} show consistent improvements on both ELVIS-S and LMOT-S. While LMOT-S scores are substantially lower than ELVIS-S and synthetic YouTube-VIS 2019, due to its highly challenging scenes (see~\cref{ssec:suppmat_lmots_limitations}), ELVIS still yields clear qualitative gains (\cref{fig:visual_comparison_lmots_frames_segms}), with fewer false positives and better tracking.

\vspace{2mm}
\noindent \textbf{Comparison against two-stage baselines.}
We compare our method against two-stage baselines to validate the necessity of an end-to-end low-light VIS method. We selected SOTA low-light enhancement methods~\cite{wang2021sdsd, zhang2021stablellve, feijoo2025DarkIR} as the pre-processing techniques and compared against GenVIS with our ELVIS framework.~\cref{tab:results_comparisons_twostage} show clear performance improvements across AP and AR metrics. 
The lower VIS performance of two-stage baselines compared to ELVIS stems from the enhancement outputs, which introduce visual artifacts and produces outputs optimized for perceptual quality rather than for preserving features relevant to semantic understanding by VIS models. This discrepancy leads to inferior VIS performances (see~\cref{fig:suppmat_two_stage_baselines}).

\begin{figure*}[ht]
\centering
\setlength{\tabcolsep}{1pt}
\renewcommand{\arraystretch}{1.0}

\begin{tabular}{@{} >{\centering\arraybackslash}m{0.3cm}
                 >{\centering\arraybackslash}m{0.18\textwidth}
                 >{\centering\arraybackslash}m{0.18\textwidth}
                 >{\centering\arraybackslash}m{0.18\textwidth}
                 >{\centering\arraybackslash}m{0.18\textwidth}
                 >{\centering\arraybackslash}m{0.18\textwidth} >
                 {\centering\arraybackslash}m{0.3cm} @{}}

     \raisebox{0.9\height}{\rotatebox[origin=c]{90}{DID~\cite{fu2023did}}} &
    \includegraphics[width=0.18\textwidth]{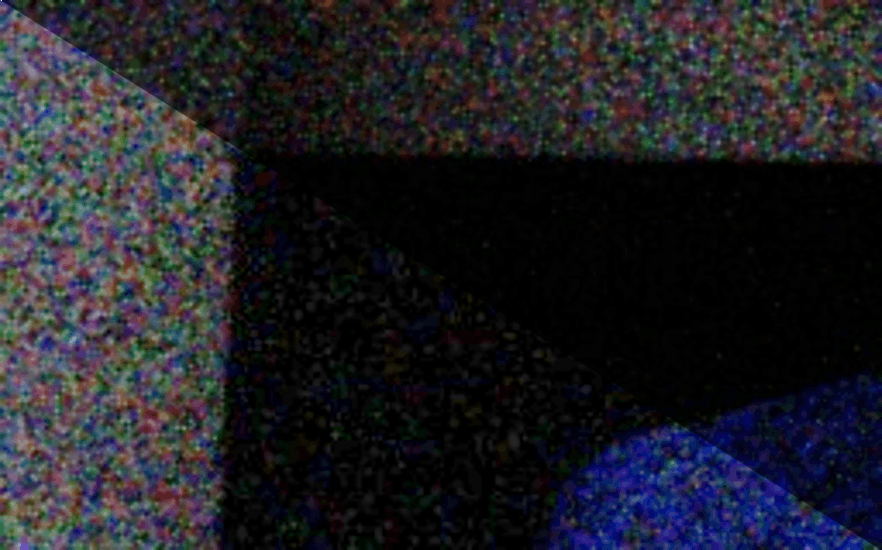} &
    \includegraphics[width=0.18\textwidth]{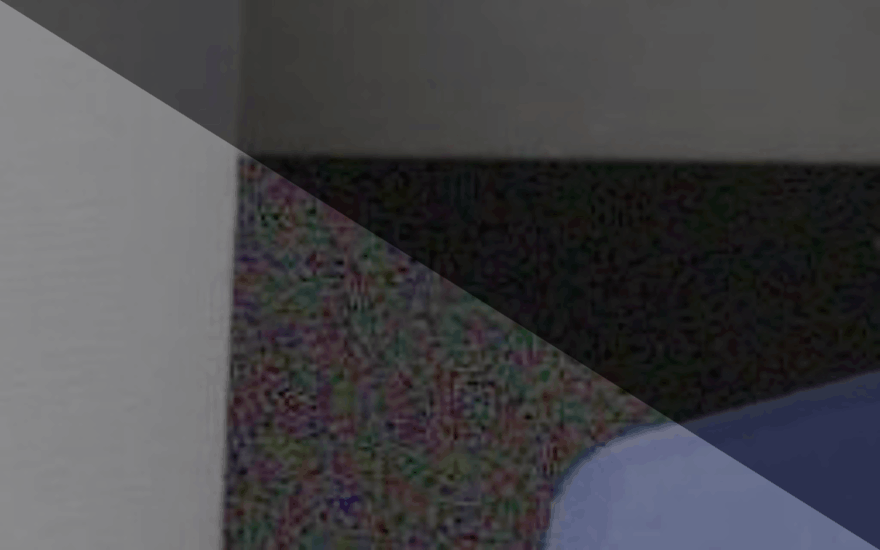} &
    \includegraphics[width=0.18\textwidth]{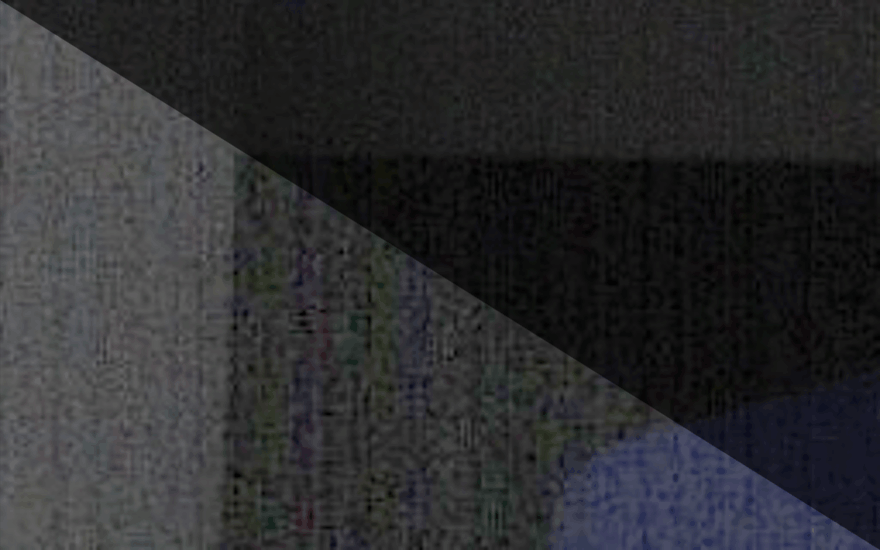} &
    \includegraphics[width=0.18\textwidth]{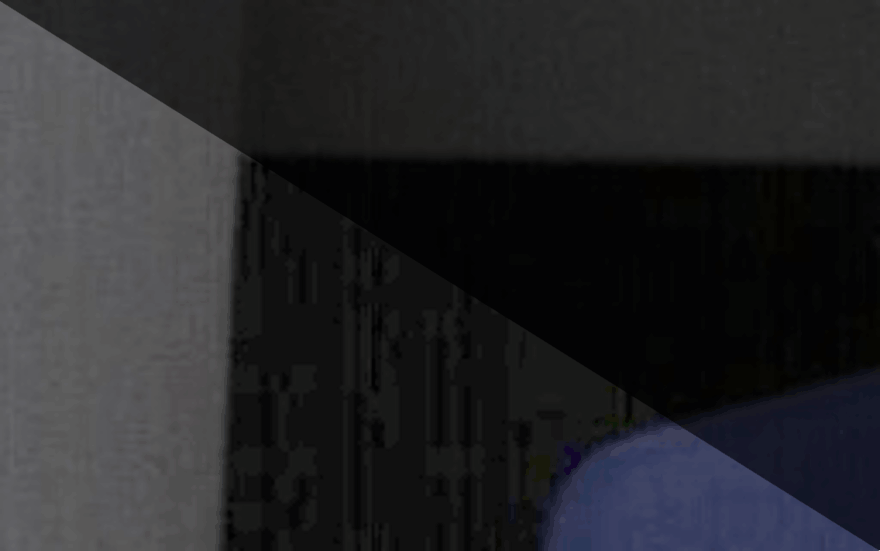} &
    \includegraphics[width=0.18\textwidth]{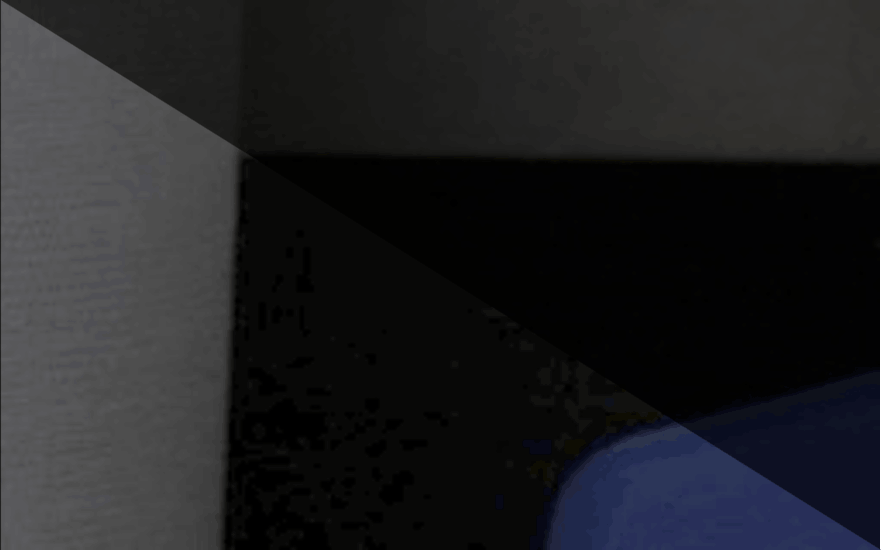} & \raisebox{0.5\height}{\rotatebox[origin=c]{270}{Low}}\\[-2pt]
    
    \raisebox{0.8\height}{\rotatebox[origin=c]{90}{SDSD~\cite{wang2021sdsd}}} &
    \includegraphics[width=0.18\textwidth]{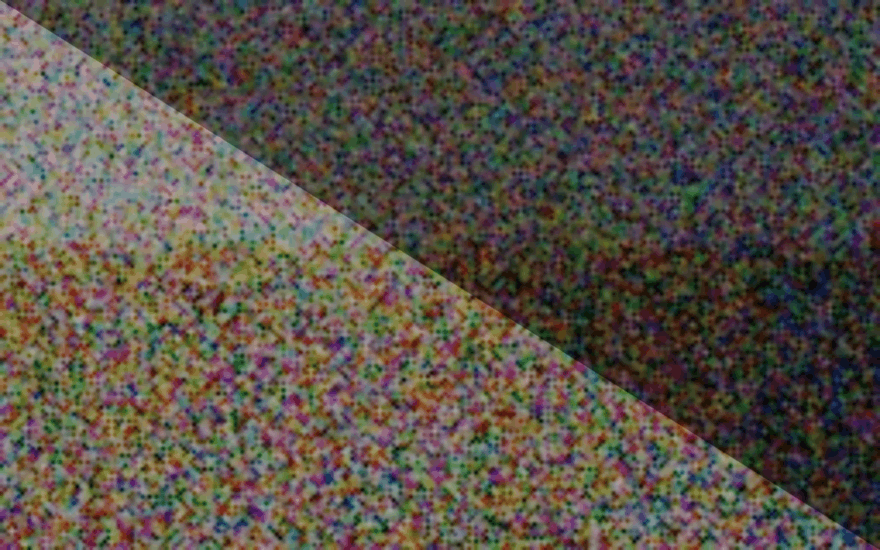} &
    \includegraphics[width=0.18\textwidth]{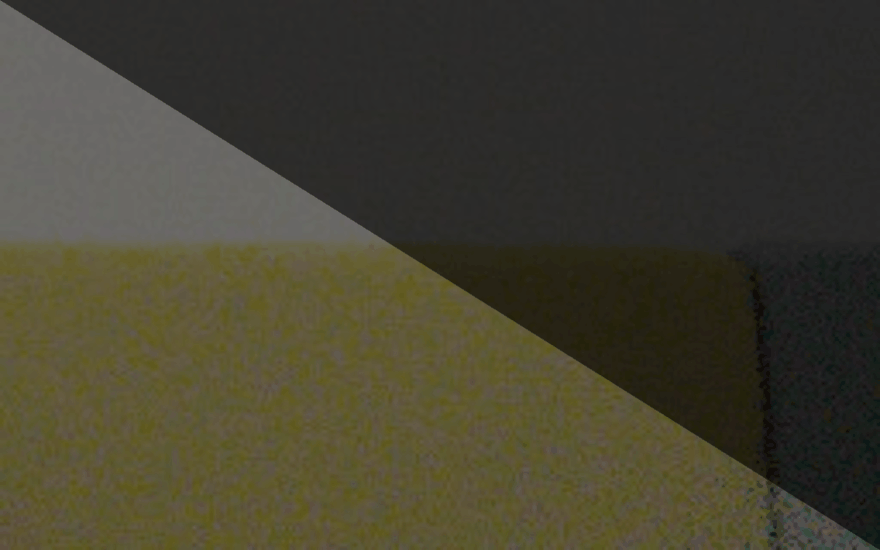} &
    \includegraphics[width=0.18\textwidth]{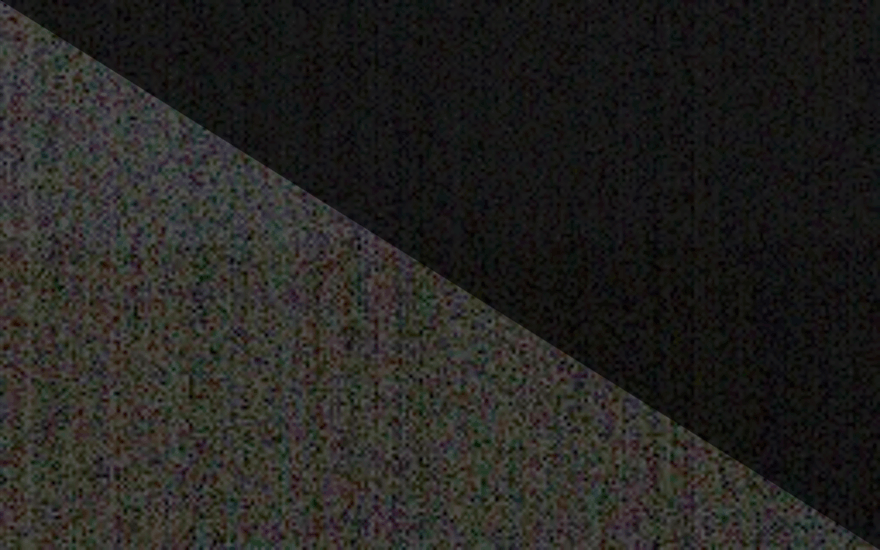} &
    \includegraphics[width=0.18\textwidth]{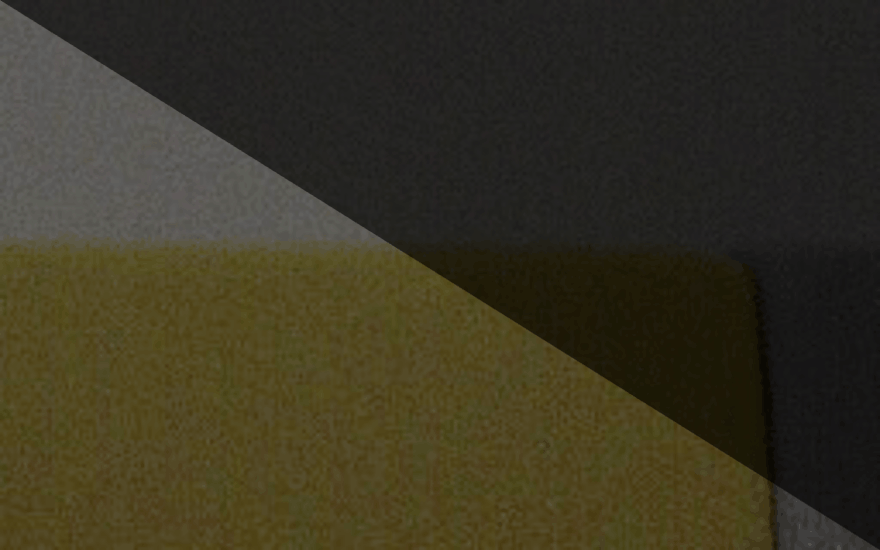} &
    \includegraphics[width=0.18\textwidth]{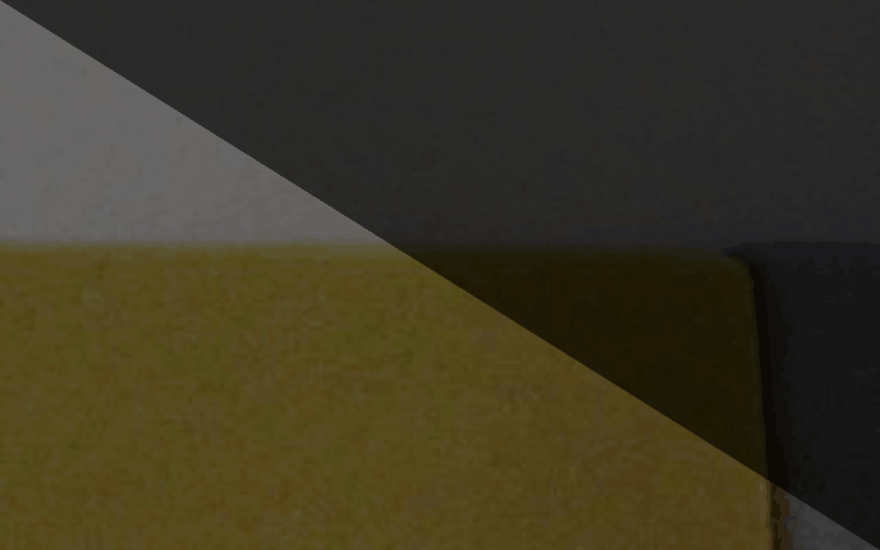} & \raisebox{0.5\height}{\rotatebox[origin=c]{270}{Medium}}\\[-2pt]

    \raisebox{0.7\height}{\rotatebox[origin=c]{90}{LMOT~\cite{wang2024lmot}}} &
    \includegraphics[width=0.18\textwidth]{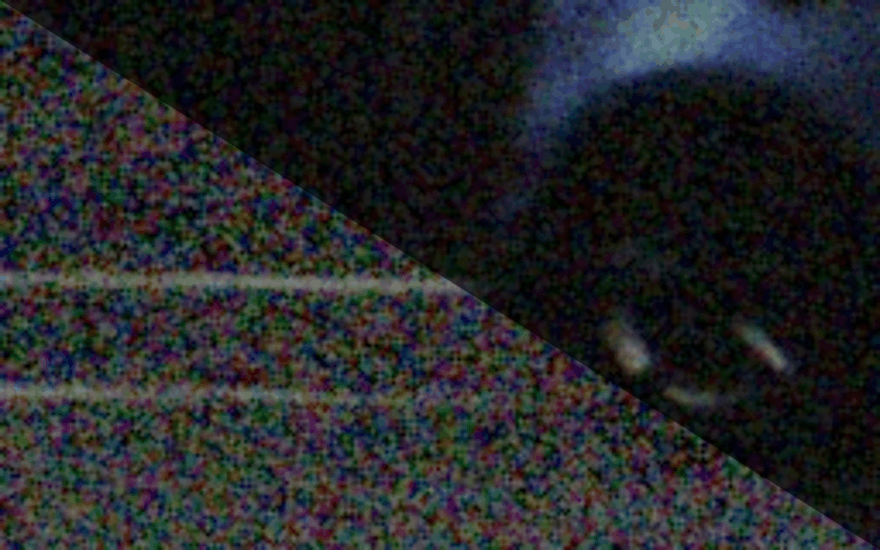} &
    \includegraphics[width=0.18\textwidth]{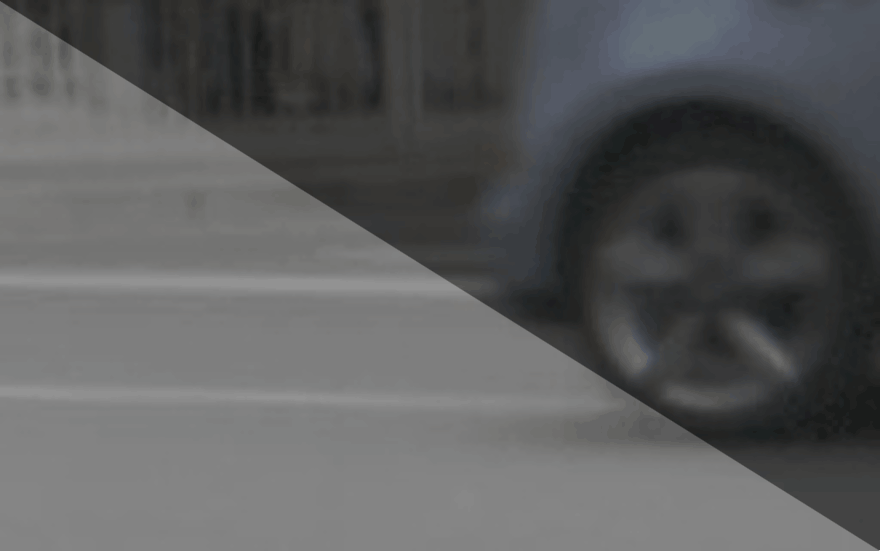} &
    \includegraphics[width=0.18\textwidth]{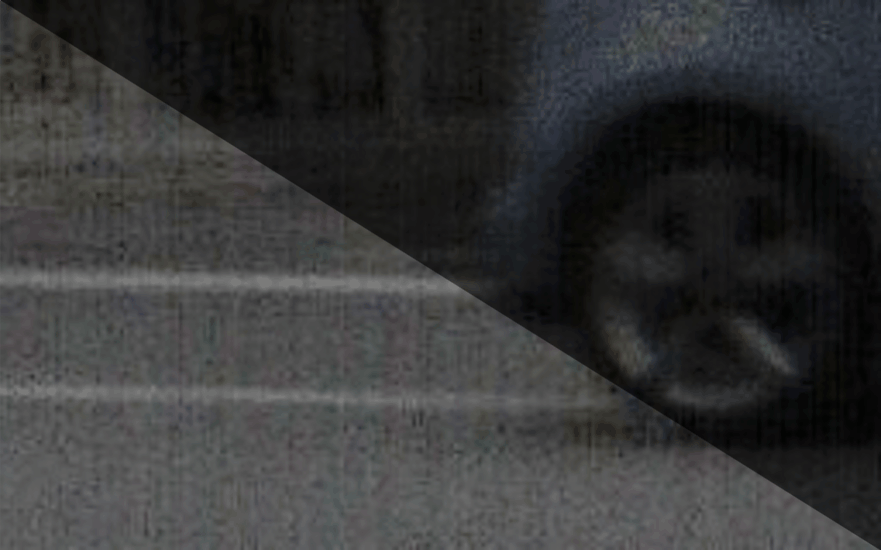} &
    \includegraphics[width=0.18\textwidth]{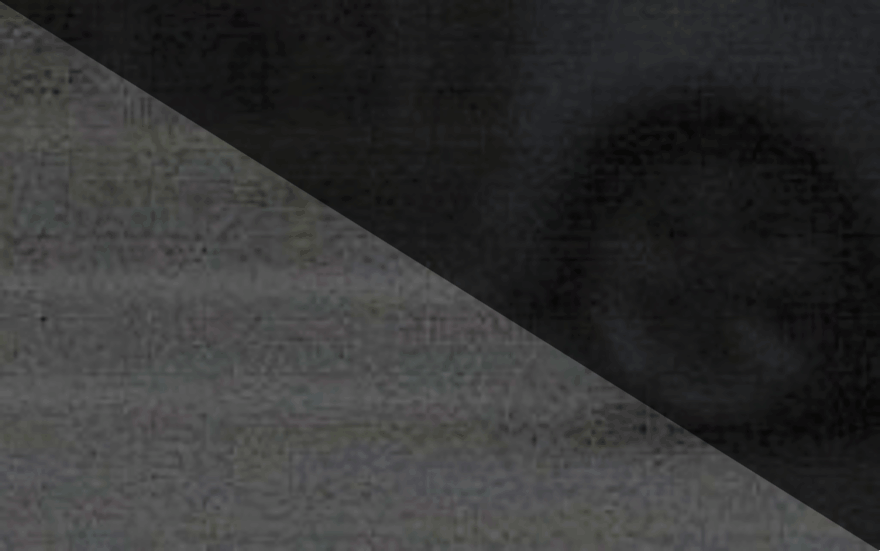} &
    \includegraphics[width=0.18\textwidth]{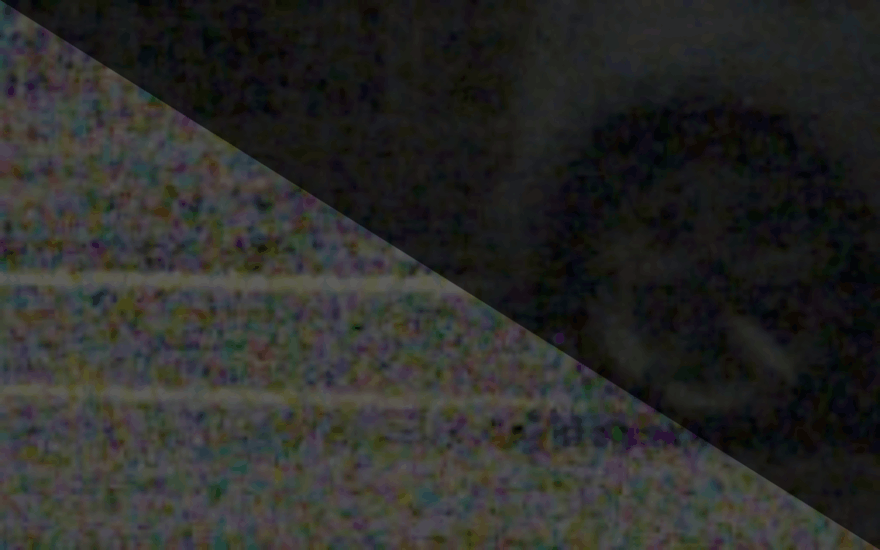} &
    \hspace{-1mm}
    \raisebox{0.5\height}{\rotatebox[origin=c]{270}{High}}\\[-2pt]

    & Lv~\etal~\cite{lv2021agllnet} & Cui~\etal~\cite{cui2021maet} & Lin~\etal~\cite{lin2025den} & Ours & Real & \\
\end{tabular}

\caption{Qualitative analysis of the several synthetic pipelines against the frames from real low-light datasets (SDSD~\cite{wang2021sdsd}, DID~\cite{fu2023did}, LMOT~\cite{wang2024lmot}). The brightness and contrast in the bottom-left triangles of each patch were adjusted by 40\% for better visibility.}
\label{fig:visual_comparison_synth_crops}
\end{figure*}

\vspace{2mm}
\noindent \textbf{Comparison against synthetic noise pipelines.}
We also compare our synthetic pipeline against other physics-based low-light pipelines~\cite{lv2021agllnet, cui2021maet, lin2025den} for training GenVIS and enabling real low-light domain adaptation, with the results shown in~\cref{tab:results_comparisons_pipelines}. We do not compare against~\cite{chen2023lis} as their synthetic pipeline produces RAW images which are not suitable for the VIS methods we employ.
Additionally, we compare our method against a variant of the synthetic pipeline with randomly sampled $\phi$ to validate the necessity of VDP-Net. Across all metrics, the pipeline with VDP-Net consistently achieves superior performance, despite not explicitly modeling spatially correlated noise as in~\cite{lv2021agllnet, cui2021maet}. This is expected, since such degradations are often camera-specific and do not generalize well. Notably, our pipeline can readily incorporate degradations attributed to camera-specific ISPs if desired. We also find that learning a structured degradation distribution with VDP-Net proves more effective than random sampling.
Examples of synthesized datasets~\cite{fu2023did, wang2021sdsd, wang2024lmot} are shown in~\cref{fig:visual_comparison_synth_crops}, demonstrating our pipeline’s ability to generate diverse degradation levels.

\subsection{Ablation Study of VDP-Net}
\cref{tab:ablation_vdpnet} presents an ablation study of various configurations of VDP-Net. For each experiment, we train the model on the synthetic YouTube-VIS 2019~\cite{yang2019ytvis19} dataset using randomly sampled $\phi$ values (see~\cref{ssec:VDPNet}) for 100 epochs with a batch size of 16. Each input video consists of 5 frames with a patch size of 256$\times$256 pixels. We set $\lambda_1$ and $\lambda_2$ to 1 when CA loss is applied.
This ablation study analyzes the effects of the temporal fusion block, and the inclusion of CA loss.
We evaluate each model on the Real-LOL-Blur~\cite{zhou2022lednet} dataset, chosen for its wide range of blur levels in low-light frames, allowing us to validate the model’s capability to handle temporal degradations.
To quantify the distributional differences between synthesized and real low-light frames, we use the Kullback–Leibler Divergence (KLD)~\cite{monakhova2022starlight,Zhang2023RAWnoise,lin2025den}. We also evaluate using the Fréchet Inception Distance (FID)~\cite{heusel2017fid}; a standard metric for measuring the similarity between generated and real image distributions.
The results confirm both VDP-Net components are essential, consistently improving KLD and FID scores.

\begin{table}[!t]
    \centering
    \small
    \caption{Ablation of VDP-Net components, evaluating similarity of synthetic outputs to Real-LOL-Blur~\cite{zhou2022lednet} via KLD and FID~\cite{heusel2017fid}.}
    \begin{tabular}{cc|cc}
        \toprule
        CA Loss & Temporal Fusion & KLD ($\downarrow$) & FID ($\downarrow$) \\
        \midrule
        \xmark & \xmark & 0.573 & 85.369 \\
        \cmark & \xmark & 0.485 & 91.100 \\
        \cmark & \cmark & \textbf{0.469} & \textbf{85.156 }\\
        \bottomrule
    \end{tabular}
    \vspace{-3mm}
    \label{tab:ablation_vdpnet}
\end{table}

\subsection{Low-Light Video Enhancement (LLVE)}
We further analyze the effectiveness of our synthetic pipeline by comparing it with other synthetic pipelines on LLVE tasks, using SDSD-Net~\cite{wang2021sdsd} as the enhancement model. For training, we synthesize the input videos by applying the physics-based synthetic pipelines~\cite{lv2021agllnet, cui2021maet, lin2025den} to the normal-light ground truths from two LLVE datasets (SDSD~\cite{wang2021sdsd} and DID~\cite{fu2023did}) for 120k iterations, and evaluate on the corresponding real low-light test sets.
We evaluate using standard enhancement metrics: Peak Signal-to-Noise Ratio (PSNR), Structural Similarity Index Measure (SSIM) and Learned Perceptual Image Patch Similarity (LPIPS)~\cite{zhang2018lpips}.~\cref{tab:results_pipelines_llve} shows our method achieves the best performance across all metrics on DID, and ranks best in LPIPS and second in PSNR/SSIM on SDSD; confirming that our pipeline can synthesize realistic low-light degradations.
Further LLVE analysis is provided in~\cref{sec:suppmat_llve}.

\begin{table}[!t]
 \centering
  \caption{Evaluation of training SDSD-Net~\cite{wang2021sdsd} with physics-based low-light synthetic pipelines across SDSD~\cite{wang2021sdsd} and DID~\cite{fu2023did} using PSNR ($\uparrow$), SSIM ($\uparrow$) and LPIPS ($\downarrow$). \textbf{Bold} and \underline{underlined} denote the best and second-best performances respectively.}
    \setlength{\tabcolsep}{3pt}
    \resizebox{1.0\linewidth}{!}{\begin{tabular}{l|ccc|ccc}
    \toprule
    Method & \multicolumn{3}{c|}{SDSD} & \multicolumn{3}{c}{DID}\\
    & PSNR & SSIM & LPIPS & PSNR & SSIM & LPIPS \\
    \midrule
    Lv \etal~\cite{lv2021agllnet} & 15.249 & 0.399 & 0.448 & 12.185 & \underline{0.580} & 0.315 \\
    Cui \etal~\cite{cui2021maet} & 15.684 & 0.448 & 0.303 & 12.220 & 0.476 & \underline{0.256} \\
    Lin \etal~\cite{lin2025den} & \textbf{21.970} & \textbf{0.664} & \underline{0.285} & \underline{13.102} & 0.531 & 0.351 \\
    Ours & \underline{16.946} & \underline{0.611} & \textbf{0.198} & \textbf{14.266} & \textbf{0.676} & \textbf{0.164} \\ 
    \bottomrule
    \end{tabular}}
    \vspace{-2mm}
    \label{tab:results_pipelines_llve}
\end{table}
\section{Conclusion}


In this paper we propose ELVIS, the first domain adaptation framework to improve video instance segmentation in the dark. ELVIS provides the following: 1) an unsupervised synthetic video-specific pipeline (which includes blur degradations), 2) a degradation profile estimation network (VDP-Net) for calibration-free synthesis, and 3) the addition of the enhancement decoder module into networks to disentangle degradation features from the underlying content. We show that implementing ELVIS into existing SOTA methods can easily boost low-light VIS performances. While our real low-light evaluation is preliminary, we identify the release of properly annotated low-light VIS benchmarks as critical future work for the community.

\section*{Acknowledgements}
This work was supported by the UKRI MyWorld Strength in Places Programme (SIPF00006/1), EPSRC Doctoral Training Partnerships (EP/W524414/1) and the University of Bristol.

{
    \small
    \bibliographystyle{ieeenat_fullname}
    \bibliography{main}
}

\clearpage
\setcounter{page}{1}
\maketitlesupplementary

\setcounter{figure}{0}
\renewcommand{\thefigure}{S\arabic{figure}}
\setcounter{table}{0}
\renewcommand{\thetable}{S\arabic{table}}
\setcounter{section}{0}
\renewcommand{\thesection}{S\arabic{section}}

\begin{strip}
    \centering
    \vspace{-0.75cm}
    \includegraphics[width=\textwidth]{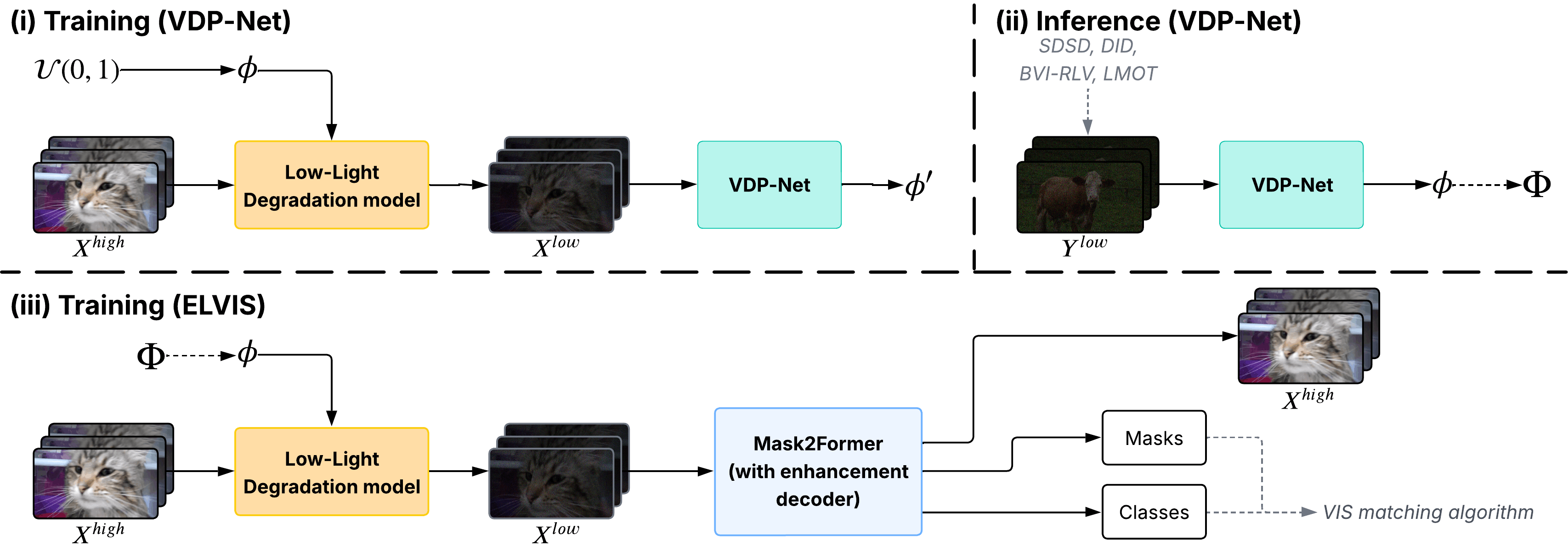}
    \captionof{figure}{Step-by-step process for using the ELVIS framework: (i) train VDP-Net to predict degradation profile $\phi$, (ii) build $\Phi$ from real low-light datasets~\cite{wang2021sdsd, fu2023did, lin2024bvi, wang2024lmot}, and (iii) use degradation profiles from $\Phi$ to create synthetic data to train ELVIS.}
    \label{fig:suppmat_diagram}
   \vspace{-0.2cm}
\end{strip}

\section{ELVIS Details}
\label{sec:suppmat_details}
\cref{fig:suppmat_diagram} shows a step-by-step visualization of each process required for the ELVIS framework for clarity. Refer to the main paper for the implementation details of each step.


\section{LMOT-S Benchmark}
\subsection{Construction details}
\label{ssec:suppmat_lmots_construction}
In order to construct a large low-light VIS benchmark for evaluation, we had to ensure that the Segment Anything Model (SAM)~\cite{kirillov2023segmentanything} was capable of producing high-quality segmentation masks for the dataset. LMOT~\cite{wang2024lmot} is well-suited to our requirements as the dataset provides both video object tracking annotations \textit{and} paired low-light and normal-light frames. The paired normal-light frames allows us to collect segmentation masks without the performance impact arising from low-light degradations, while the video object tracking annotations allows us to easily prompt Segment Anything to produce segmentation masks for the object of interest.

We construct LMOT-S by downsampling the low-light sRGB frames from the original LMOT~\cite{wang2024lmot} validation set to 720p resolution and dividing the videos in clips of 51 frames each (22 even segments per video). This enables us to evaluate on 88 videos; allowing us to determine the robustness of our method across a large scale. We then leverage the provided bounding box annotations, to feed into SAM~\cite{kirillov2023segmentanything} along with the paired normal-light frames to acquire pseudo ground truth segmentation annotations.
Finally we process the segmentation results with the original annotations to match with the YouTube-VIS 2019~\cite{yang2019ytvis19} annotation format, allowing for easy evaluation. Due to the mismatch in categories, we rename all `car' instances in LMOT-S to `sedan', rename all `motorcycle' instances to `motorbike' and remove all `bicycle' and `bus' instances. The IDs of each category are re-mapped to align with the IDs from YouTube-VIS 2019.


\begin{table}[!t]
    \centering
    \caption{Statistics of count of instances from YouTube-VIS~\cite{yang2019ytvis19} versus LMOT-S validation sets.}
    \begin{tabular}{l|ccc}
    \toprule
    Dataset & Min & Max & Mean \\
    \midrule
    YouTube-VIS 2019~\cite{yang2019ytvis19} & 1 & 6 & 1.7 \\
    LMOT-S & 9 & 75 & 31.5 \\
    \bottomrule
    \end{tabular}
    \label{tab:suppmat_segm_stats}
\end{table}

\begin{table}[!t]
 \centering
 \caption{LMOT-S results on GenVIS~\cite{heo2023genvis} (ResNet-50) trained on our synthetic low-light YouTube-VIS 2019 videos. Original pretrained normal-light results are shown for comparison.}
    \resizebox{1.0\linewidth}{!}{\begin{tabular}{l|c|cccccc}
    \toprule
    Method & Light & AP & AP$_{50}$ & AP$_{75}$ & AR$_{1}$ & AR$_{10}$ & AR$_{100}$ \\
    \midrule
    GenVIS~\cite{heo2023genvis} & Low & 6.6 & 14.5 & 5.2 & 5.4 & 9.8 & 11.7 \\
    GenVIS~\cite{heo2023genvis} + ELVIS & Low & 6.7 & 15.5 & 4.4 & 5.3 & 12.1 & 14.0 \\
    \midrule
    GenVIS~\cite{heo2023genvis} (pre-trained) & Normal & 11.7 & 21.4 & 10.5 & 8.8 & 16.8 & 21.0 \\
    
    \bottomrule
    \end{tabular}}
 
 \label{tab:suppmat_lmots_genvis_all}
\end{table}

\begin{table}[t]
    \centering
    \caption{Further evaluation on LMOT-S for two-stage baselines and synthetic pipelines. \textbf{Bold} indicates the best performance.}
    \small
    
    \begin{minipage}{0.48\linewidth}
        \centering
        \begin{tabular}{l|c}
            \toprule
            \multicolumn{2}{c}{Two-Stage Baselines} \\
            \midrule
            Method & AR$_{100}$ \\
            \midrule
            SDSD-Net~\cite{wang2021sdsd} & 5.4 \\
            StableLLVE~\cite{zhang2021stablellve} & 8.1 \\
            DarkIR~\cite{feijoo2025DarkIR} & 8.0 \\
            ELVIS & \textbf{14.0} \\
            \bottomrule
        \end{tabular}
    \end{minipage}
    \hfill
    \begin{minipage}{0.48\linewidth}
        \centering
        \begin{tabular}{l|c}
            \toprule
            \multicolumn{2}{c}{Synthetic Pipelines} \\
            \midrule
            Method & AR$_{100}$ \\
            \midrule
            Lv \etal~\cite{lv2021agllnet} & 9.5 \\
            Cui \etal~\cite{cui2021maet} & 10.2 \\
            Lin \etal~\cite{lin2025den} & 9.5 \\
            Ours (random) & 7.4 \\
            Ours & \textbf{11.7} \\
            \bottomrule
        \end{tabular}
    \end{minipage}

    \vspace{-5pt}
    \label{tab:suppmat_lmots_ar100}
\end{table}

\subsection{Extra results}
As we had increased the number of detections to 100, the AR$_{100}$ becomes a meaningful metric for understanding a method's performance at detecting and segmenting a large number of instances. We report these results in~\cref{tab:suppmat_lmots_ar100} which shows our method also has the best AR$_{100}$ scores.

\subsection{Limitations of LMOT-S benchmark}
\label{ssec:suppmat_lmots_limitations}

\begin{figure*}[!ht]
    \centering
    \includegraphics[width=0.82\textwidth]{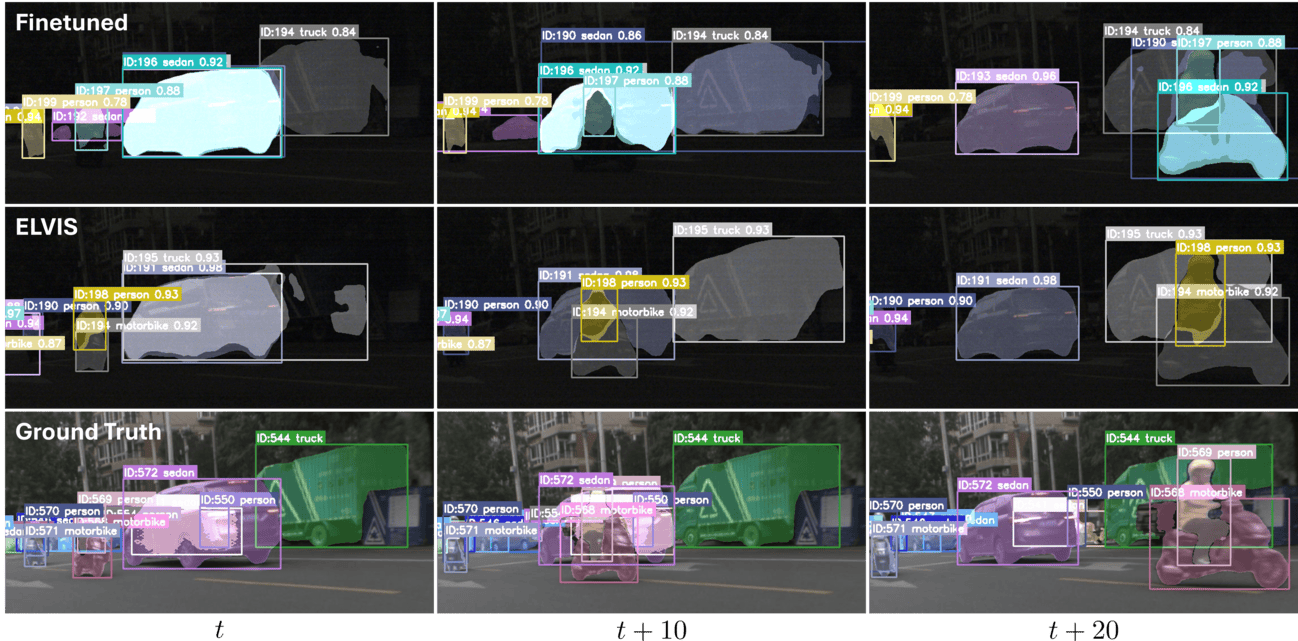}
    \caption{Visual comparison of segmentation and tracking success and failure cases on the LMOT-S dataset using GenVIS~\cite{heo2023genvis} method with a ResNet-50 backbone finetuned on our synthetic data (top row) versus implementing our ELVIS framework (middle row), with ground truth (bottom row) for reference. The columns represent frames in the example video, sampled every 10 frames from time $t$, to show the tracking performances.}
    \label{fig:suppmat_lmots_segm}
\end{figure*}

Although LMOT-S~\cite{wang2024lmot} offers a useful preliminary evaluation for real-world settings, it is a far more challenging benchmark than YouTube-VIS 2019~\cite{yang2019ytvis19}.
One main reason for the notably lower performances when evaluating on the LMOT-S dataset can be attributed to the high instance counts (see ~\cref{tab:suppmat_segm_stats}). LMOT-S contains at least 9 instances per video, whereas the YouTube-VIS 2019 validation set contains at most 6; existing VIS methods are not designed to handle such large number of instances.

Another challenge is the cross-over issue, which leads to a high number of identity switches. Combined with the fact that these methods are designed to track fewer instances, this results in frequent identity switches and a failure to form new tracklets. An example can seen in~\cref{fig:suppmat_lmots_segm}, where in the finetuned GenVIS~\cite{heo2023genvis} model, it transferred the sedan identity (ID: 196) onto the motorcycle as it drove past the car, and created a new identity (ID: 193) for the car. ELVIS reduces the occurrences of such issues, but they still persist (see the truck in frame $t$). In~\cref{tab:suppmat_lmots_genvis_all}, we also compare against GenVIS with off-the-shelf weights provided by the authors, on the paired normal-light frames of LMOT-S, to emphasize the limitations of the benchmark.

As such, we encourage considering results on both ELVIS-S and LMOT-S to obtain a more comprehensive understanding of real low-light performance.

\begin{table}[!t]
 \centering
  \caption{Evaluation of SDSD-Net~\cite{wang2021sdsd} trained with physics-based low-light synthetic pipelines across SDSD~\cite{wang2021sdsd} and DID~\cite{fu2023did} using PSNR ($\uparrow$), SSIM ($\uparrow$), and LPIPS ($\downarrow$). Outputs are histogram-matched. \textbf{Bold} indicates the best performances.}
    \resizebox{1.0\linewidth}{!}{\begin{tabular}{l|ccc|ccc}
    \toprule
    Method & \multicolumn{3}{c|}{SDSD} & \multicolumn{3}{c}{DID}\\
       & PSNR & SSIM & LPIPS & PSNR & SSIM & LPIPS \\
    \midrule
     Lv \etal~\cite{lv2021agllnet} & 21.421 & 0.464 & 0.563 & 16.606 & 0.846 & 0.476 \\
     Cui \etal~\cite{cui2021maet} & 19.199 & 0.645 & 0.380 & 22.042 & 0.824 & 0.220 \\
     Lin \etal~\cite{lin2025den} & \textbf{23.088} & 0.645 & 0.393 & 17.354 & 0.624 & 0.450 \\
     Ours & 20.703 & \textbf{0.688} & \textbf{0.282} & \textbf{22.794} & \textbf{0.846} & \textbf{0.197} \\ 
     \bottomrule
    \end{tabular}}
   \vspace{-2mm}
    \label{tab:suppmat_llve_histmatch}
\end{table}


\begin{figure}[t]
    \centering
    \begin{subfigure}{0.49\linewidth}
        \centering
        \includegraphics[width=\linewidth]{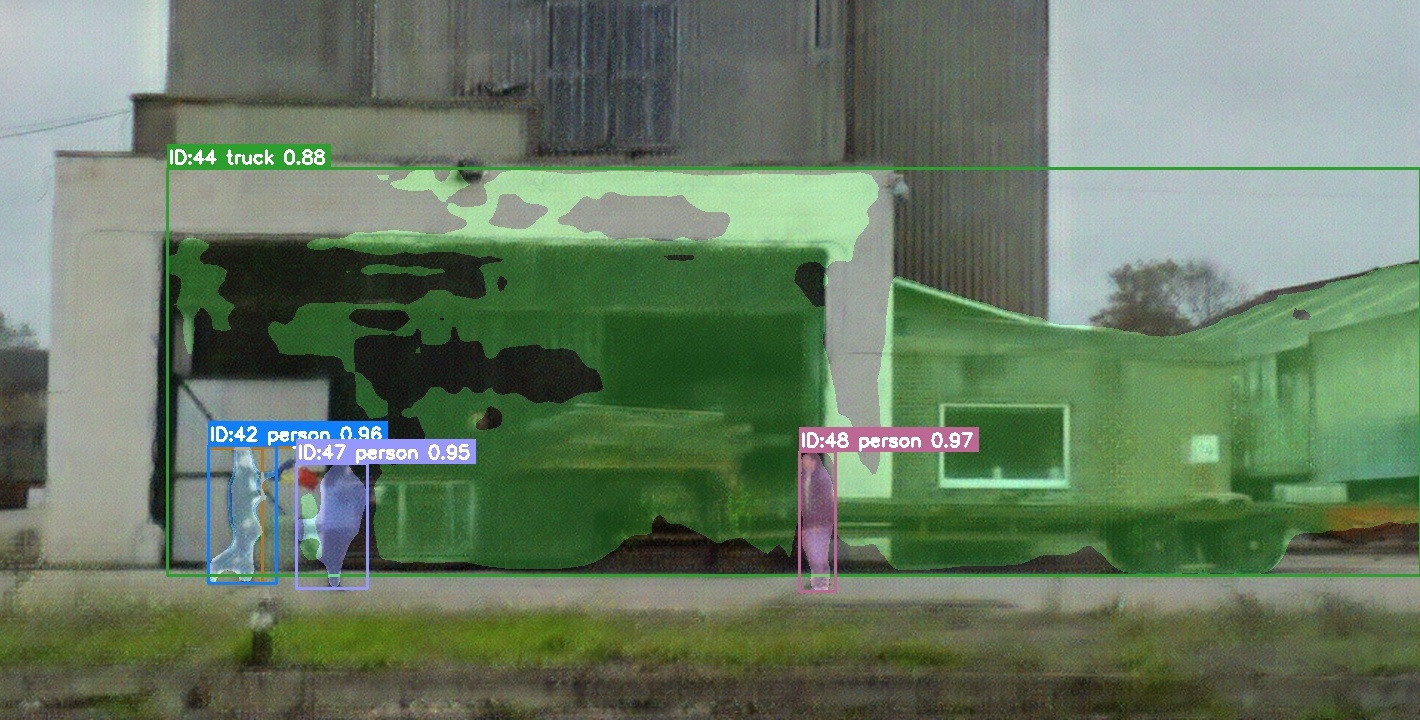}
        \caption{SDSDNet~\cite{wang2021sdsd}}
    \end{subfigure}
    \hfill
    \begin{subfigure}{0.49\linewidth}
        \centering
        \includegraphics[width=\linewidth]{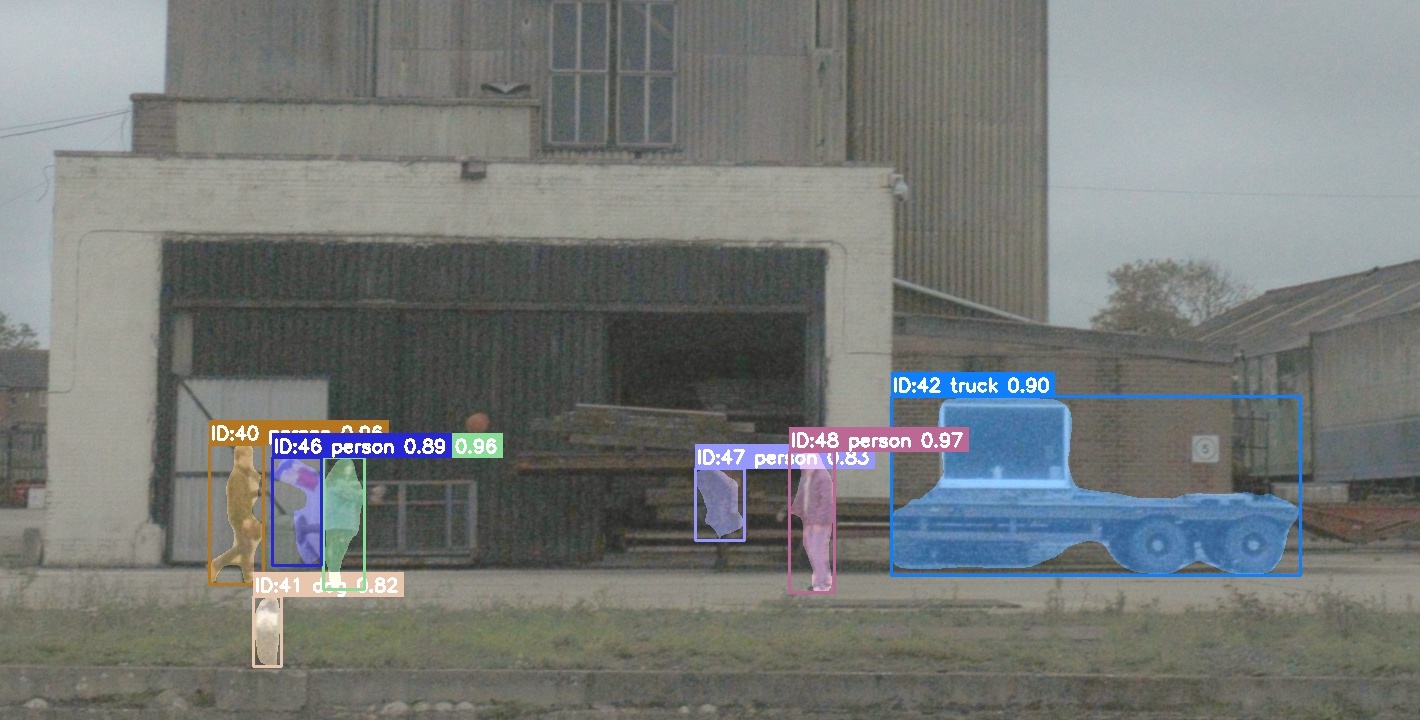}
        \caption{StableLLVE~\cite{zhang2021stablellve}}
    \end{subfigure}

    \vspace{0.2em}

    \begin{subfigure}{0.49\linewidth}
        \centering
        \includegraphics[width=\linewidth]{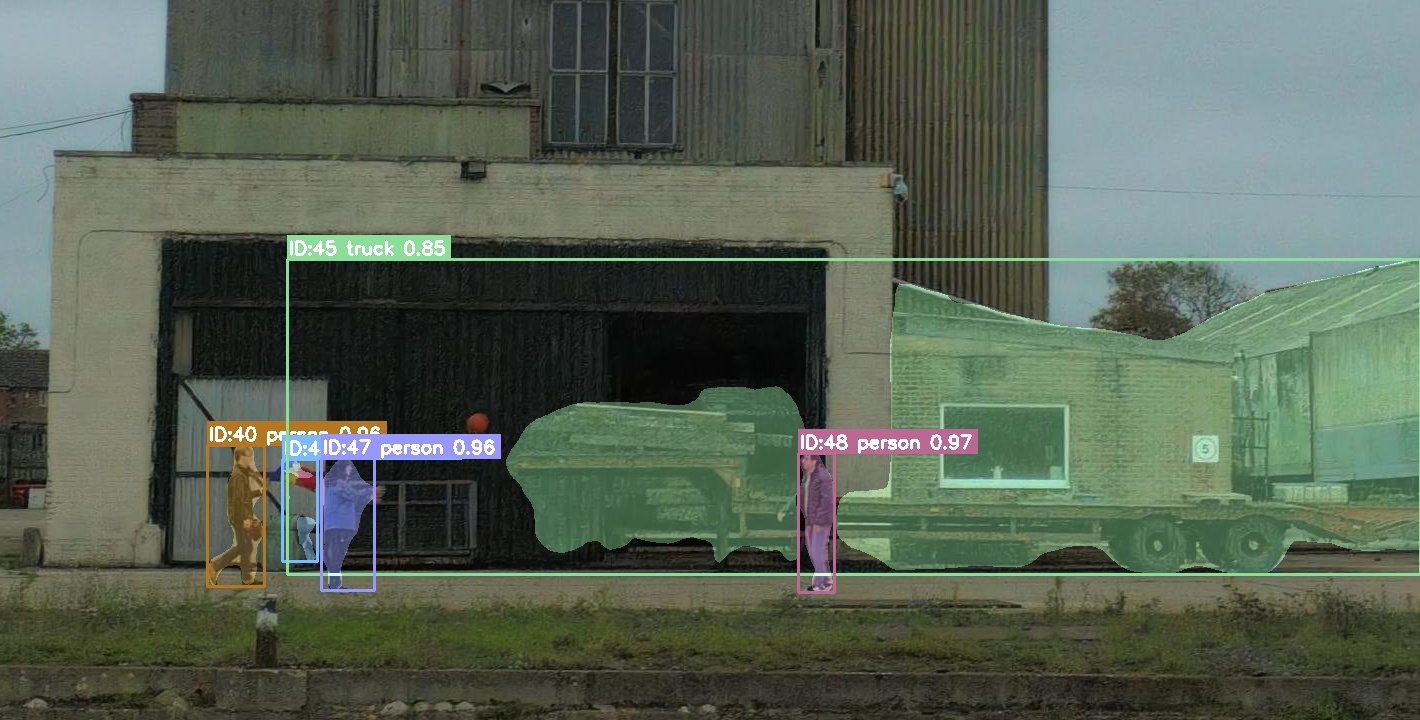}
        \caption{DarkIR~\cite{feijoo2025DarkIR}}
    \end{subfigure}
    \hfill
    \begin{subfigure}{0.49\linewidth}
        \centering
        \includegraphics[width=\linewidth]{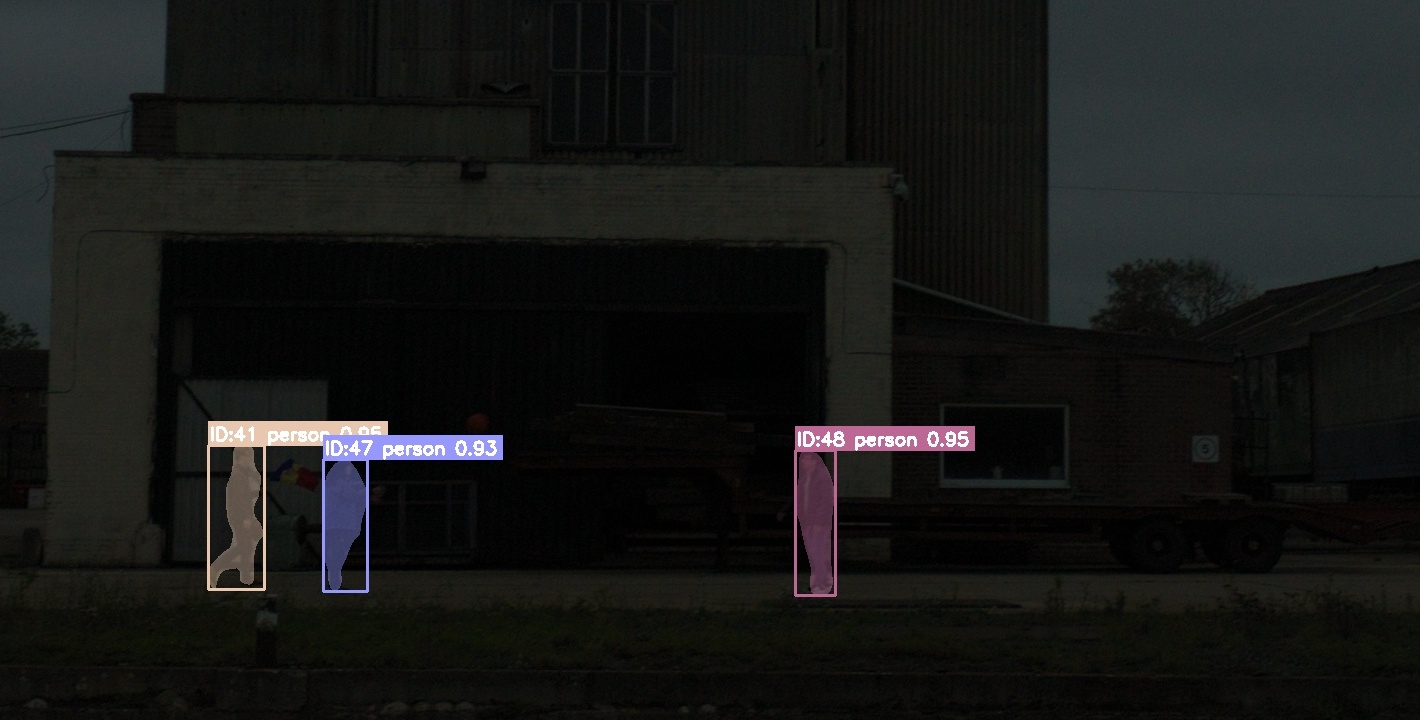}
        \caption{Ours}
    \end{subfigure}

    \caption{Visual comparison of VIS predictions using two-stage baselines versus ELVIS on ELVIS-S, with the enhanced outputs.}
    \label{fig:suppmat_two_stage_baselines}
\end{figure}

\section{Two-Stage Baseline Outputs}
\cref{fig:suppmat_two_stage_baselines} shows a visual example of how the enhancement outputs of~\cite{wang2021sdsd, zhang2021stablellve, feijoo2025DarkIR} resulted in lower VIS performances. We find that the enhanced outputs produced many false positive detections. Despite our method being unsuccessful in detecting the truck, all of the two-stage baselines incorrectly segmented the buildings as part of the truck. SDSD-Net~\cite{wang2021sdsd} performed the worst consistently, likely attributing to the artifacts it frequently produces, and heavily oversegmented for the truck. StableLLVE~\cite{zhang2021stablellve} outputted a perceptually realistic output but when passed into GenVIS R50, resulted in incorrect detections such as a dog and extra people. DarkIR~\cite{feijoo2025DarkIR} produced the highest quality enhanced output but still incorrectly classified the building as part of the truck. We would like to highlight that while DarkIR generally produced the perceptually most accurate outputs, it underperforms StableLLVE on both ELVIS-S and LMOT-S (\cref{tab:results_comparisons_twostage}), suggesting that its enhancement process degrades or suppresses semantic features necessary for downstream video instance segmentation.
Despite being the prevailing paradigm for low-light downstream tasks, two-stage baselines exhibit clear limitations, indicating the need for further research into LLVE methods that also consider the semantic features of the scene.


\begin{figure}[!ht]
    \centering
    \includegraphics[width=\linewidth]{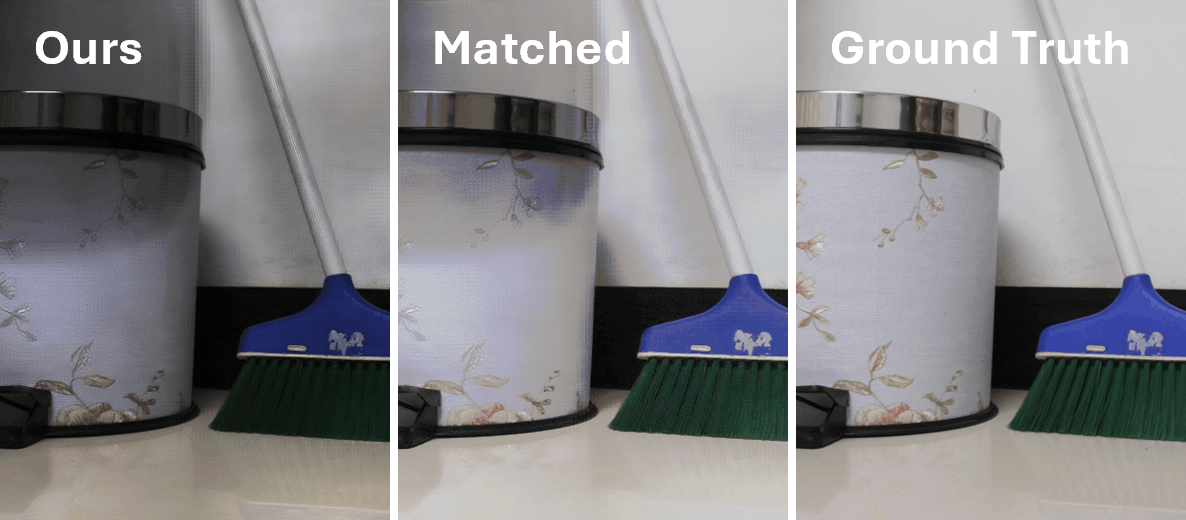}
    \caption{Example of artifacts in the enhanced output which is not removed with histogram matching. (Left) the enhanced output from SDSD-Net~\cite{wang2021sdsd} trained on our synthetic pipeline. (Center) the histogram-matched output to the ground truth. (Right) is the ground truth normal-exposed frame.}
    \label{fig:suppmat_llve_artifacts}
\end{figure}

\begin{figure}[!ht]
\centering
\setlength{\tabcolsep}{2pt}
\renewcommand{\arraystretch}{1.0}

\begin{tabular}{
    >{\centering\arraybackslash}m{0.3cm}
    >{\centering\arraybackslash}m{0.45\linewidth}
    >{\centering\arraybackslash}m{0.45\linewidth}
}

    & DID~\cite{fu2023did} & SDSD~\cite{wang2021sdsd} \\

    
    \raisebox{0.8\height}{\rotatebox[origin=c]{90}{Lv~\etal~\cite{lv2021agllnet}}} &
    \includegraphics[width=\linewidth]{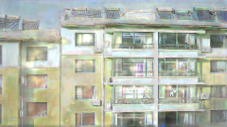} &
    \includegraphics[width=\linewidth]{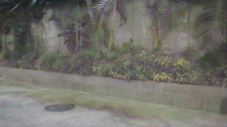} \\[-2pt]

    \raisebox{0.8\height}{\rotatebox[origin=c]{90}{Cui~\etal~\cite{cui2021maet}}} & 
    \includegraphics[width=\linewidth]{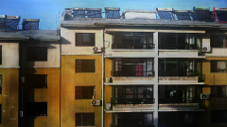} &
    \includegraphics[width=\linewidth]{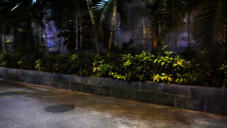} \\[-2pt]

    \raisebox{0.8\height}{\rotatebox[origin=c]{90}{Lin~\etal~\cite{lin2025den}}} &
    \includegraphics[width=\linewidth]{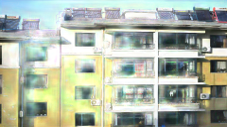} &
    \includegraphics[width=\linewidth]{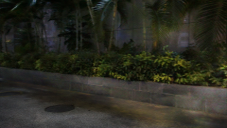} \\[-2pt]

    \raisebox{0.5\height}{\rotatebox[origin=c]{90}{Ours}} &
    \includegraphics[width=\linewidth]{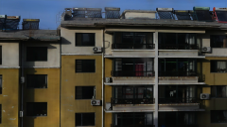} &
    \includegraphics[width=\linewidth]{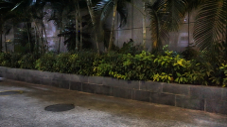} \\[-2pt]
    
    \raisebox{0.5\height}{\rotatebox[origin=c]{90}{Real}} &
    \includegraphics[width=\linewidth]{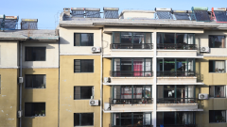} &
    \includegraphics[width=\linewidth]{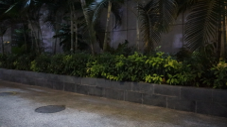} \\

\end{tabular}

\captionof{figure}{Qualitative comparison of the enhanced outputs of SDSD-Net~\cite{wang2021sdsd} trained on different synthetic pipelines from the SDSD~\cite{wang2021sdsd} and DID~\cite{fu2023did} datasets.}
\label{fig:suppmat_llve_outputs}
\end{figure}

\section{Low-Light Video Enhancement Results}
\label{sec:suppmat_llve}

We find that the outputs from SDSD-Net~\cite{wang2021sdsd} show severe artifacts (see~\cref{fig:suppmat_llve_artifacts} and~\cref{fig:suppmat_llve_outputs}); this is an observation also mentioned by Fu~\etal~\cite{fu2023did}. This explains the modest performances across all synthetic pipelines in~\cref{tab:results_pipelines_llve}.
For further analysis, we conduct additional LLVE experiments focusing on the denoising component of SDSD-Net by using different synthetic training data. Before computing the evaluation metrics, we apply histogram matching between the enhanced outputs and the ground truths. The results reflect the noise synthesis accuracy of each method.
The results are shown in~\cref{tab:suppmat_llve_histmatch}.
While PSNR and SSIM show noticeable improvements, we find that LPIPS scores worsen with histogram matching, likely because it can amplify the artifacts, reducing perceptual similarity.
We would like to mention that histogram matching is not the perfect solution for alleviating the artifacts in evaluation (see~\cref{fig:suppmat_llve_artifacts}).

\end{document}